\documentclass[runningheads]{llncs}

\usepackage{eccv}

\usepackage{eccvabbrv}

\usepackage{graphicx}
\usepackage{booktabs}

\usepackage[accsupp]{axessibility}  

\usepackage[dvipsnames]{xcolor}

\usepackage{graphicx}
\usepackage{amsmath}
\usepackage{amssymb}
\usepackage{booktabs}

\usepackage{mathtools}

\usepackage{multirow}

\usepackage[dvipsnames]{xcolor}
\usepackage{booktabs}
\usepackage{tabularx}
\usepackage{threeparttable}
\usepackage{varwidth}
\usepackage{textcomp} 

\usepackage{floatrow}
\newfloatcommand{capbtabbox}{table}[][\FBwidth]
\usepackage{pifont}
\newcommand{\cmark}{\ding{51}}%
\usepackage{dsfont}
\usepackage{bbm}

\usepackage{colortbl}

\usepackage{marvosym}

\newcommand{\tablestyle}[2]{\setlength{\tabcolsep}{#1}\renewcommand{\arraystretch}{#2}\centering\footnotesize}

\newcommand{\R}{\mathbb{R}}

\newcommand*\colourcheck[1]{%
  \expandafter\newcommand\csname #1check\endcsname{\textcolor{#1}{\ding{51}}}%
}
\colourcheck{ForestGreen}

\newcommand*\colourx[1]{%
  \expandafter\newcommand\csname #1x\endcsname{\textcolor{#1}{\ding{55}}}%
}
\colourx{blue}
\colourx{green}
\colourx{red}

\newcommand{\ours}{{BAM-DETR}}
\newcommand{\emailhref}[3][black]{\href{#2}{\color{#1}{#3}}}%

\definecolor{defaultcolor}{gray}{0.9}

\newcount\bsubfloatcount
\newtoks\bsubfloattoks
\newdimen\bsubfloatht

\makeatletter
\newcommand{\bsubfloat}[2][]{%
  \sbox\z@{#2}%
  \ifdim\bsubfloatht<\ht\z@
    \bsubfloatht=\ht\z@
  \fi
  \advance\bsubfloatcount\@ne
  \@namedef{bsubfloat\romannumeral\bsubfloatcount}{%
    \subfloat[#1]{\vbox to\bsubfloatht{\hbox{#2}\vfill}}}%
}
\newcommand{\resetbsubfloat}{\bsubfloatcount\z@\bsubfloatht=\z@}

\usepackage[pagebackref,breaklinks,colorlinks,citecolor=eccvblue]{hyperref}

\usepackage{orcidlink}

\begin{document}

\title{BAM-DETR: Boundary-Aligned Moment Detection Transformer for Temporal Sentence Grounding in Videos} 

\titlerunning{BAM-DETR}

\author{Pilhyeon Lee\inst{1}\thanks{Correspondence to: Pilhyeon Lee $<$\emailhref{mailto:pilhyeon.lee@inha.ac.kr}{\email{pilhyeon.lee@inha.ac.kr}}$>$}\orcidlink{0000-0001-6617-9080} \and
Hyeran Byun\inst{2}\orcidlink{0000-0002-3082-3214}}

\authorrunning{P.~Lee and H.~Byun}

\institute{Department of Artificial Intelligence, Inha University \and
Department of Computer Science, Yonsei University}

\maketitle

\begin{abstract}
Temporal sentence grounding aims to localize moments relevant to a language description.
Recently, DETR-like approaches achieved notable progress by predicting the center and length of a target moment.
However, they suffer from the issue of center misalignment raised by the inherent ambiguity of moment centers, leading to inaccurate predictions.
To remedy this problem, we propose a novel boundary-oriented moment formulation.
In our paradigm, the model no longer needs to find the precise center but instead suffices to predict any anchor point within the interval, from which the boundaries are directly estimated.
Based on this idea, we design a boundary-aligned moment detection transformer, equipped with a dual-pathway decoding process.
Specifically, it refines the anchor and boundaries within parallel pathways using global and boundary-focused attention, respectively.
This separate design allows the model to focus on desirable regions, enabling precise refinement of moment predictions.
Further, we propose a quality-based ranking method, ensuring that proposals with high localization qualities are prioritized over incomplete ones.
Experiments on three benchmarks validate the effectiveness of the proposed methods.
The code is available \href{https://github.com/Pilhyeon/BAM-DETR/}{here}.
  \keywords{Temporal sentence grounding \and Detection transformer}
\end{abstract}

\section{Introduction}
\label{sec:intro}

Recent years have witnessed a notable surge in the popularity of short-form video content on social media platforms like TikTok, YouTube Shorts, and Instagram Reels.
As such, users prefer to selectively engage with short \textit{moments} of interest rather than passively watch an entire long video.
This trend highlights the importance of localizing desired moments.
As a result, moment localization tasks have emerged as pivotal research topics in video understanding, including temporal action detection~\cite{zhao2017ssn,lin2018bsn,lee2023decomposed}, video summarization~\cite{zhang2016summarization,sharghi2017query-summarization}, and highlight detection~\cite{rui2000baseball-highlight,xiong2019lessismore}.
Within this context, we tackle temporal sentence grounding~\cite{anne2017mcn,zhang20202d-tan}, aiming to retrieve moments corresponding to free-form language descriptions.

To address temporal sentence grounding, numerous efforts have been undertaken in the last decade~\cite{gao2017tall,shao2018find-and-focus,yuan2019semantic-modulation,escorcia2019cal,zhang20202d-tan,liu2021context-biaffine}.
Especially, taking inspiration from DETR~\cite{carion2020detr}, query-based approaches have become a promising research direction owing to the architectural simplicity~\cite{lei2021qvhighlights,liu2022umt,moon2023qd-detr,jang2023eatr,li2023momentdiff}.
By decoding temporal spans (\ie, moments) from a handful set of learnable queries, they achieve promising grounding performance while maintaining high inference speed.

\begin{figure}[t]
    \centering
    \includegraphics[width=0.595\linewidth]{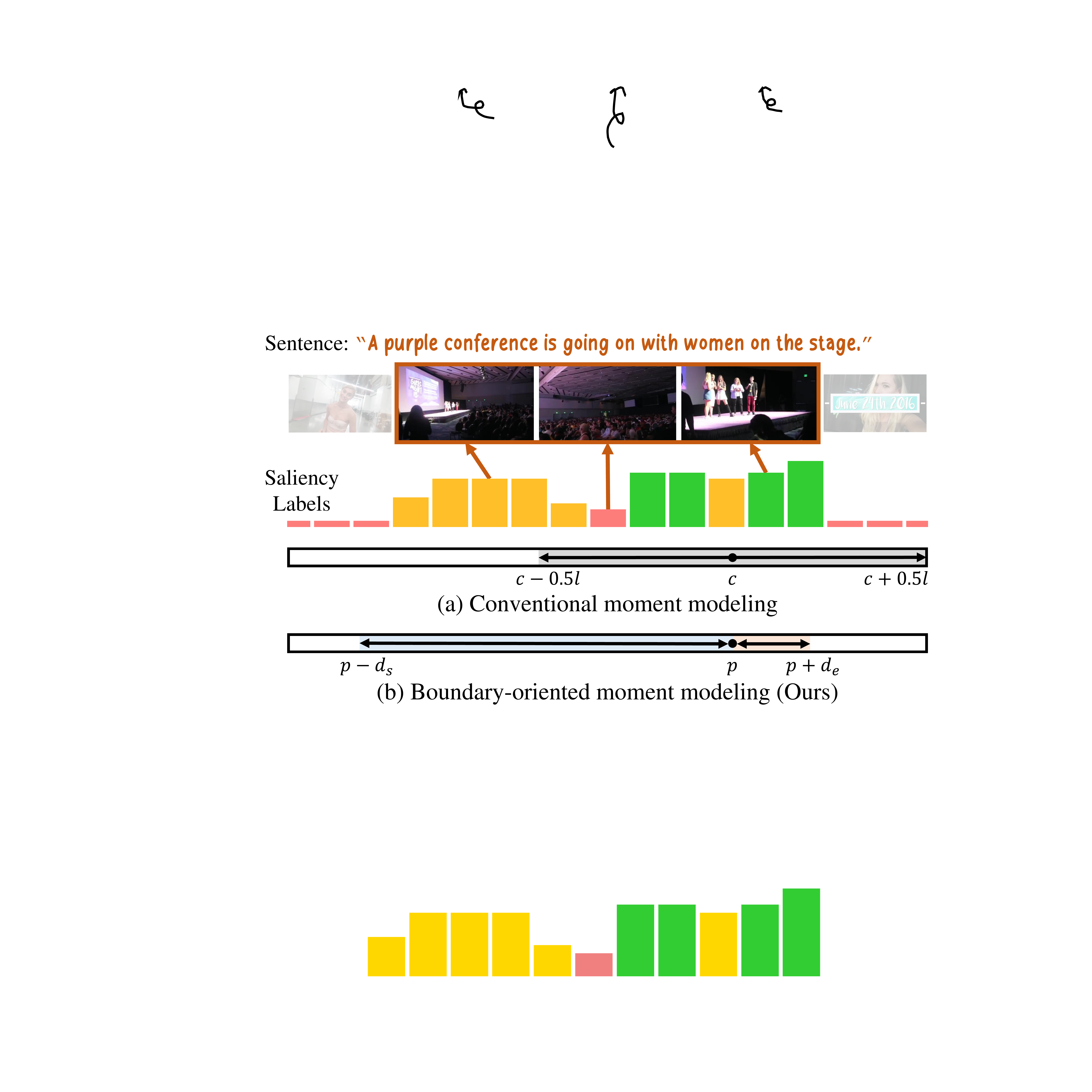}
    \caption{Comparison of moment modeling approaches under the scenario of an ambiguous center from QVHighlights. (a) The conventional method formulates a moment with a tuple of ($c$, $l$). (b) In contrast, we propose to model it with a triplet of ($p$, $d_s$, $d_e$).}
    \label{fig:center_ambiguity}
\end{figure}

Existing query-based models typically predict a moment using its center and width, \ie, ($c$, $l$), under the assumption that the boundaries are equidistant from the center.
However, such a formulation can be problematic as the center of a moment might be ambiguous.
An illustrative example is presented in \cref{fig:center_ambiguity} (top), where the frame at the center of the ground-truth moment is less relevant to the given sentence, as demonstrated by the low saliency label.
Indeed, a moment center does not necessarily serve as the best representative of the sentence. 
This ambiguity can challenge the model's ability to precisely locate the centers, and the misaligned centers lead to low-quality predictions (\cref{fig:center_ambiguity}\textcolor{red}{a}).
To probe the impact of center misalignment, we conducted a diagnostic experiment in \Cref{tab:center_misalignment}.
The results reveal that (i) existing methods struggle with detecting accurate center points and (ii) they suffer from significant performance drops when the predicted centers deviate from the ground-truth centers (\ie, large center errors).

To address this challenge, we present a novel boundary-oriented formulation for moments, as illustrated in \cref{fig:center_ambiguity}\textcolor{red}{b}, where each moment is represented by a triplet consisting of an anchor point and its distances to the boundaries, \ie, $(p, d_s, d_e)$.
This asymmetric formulation liberates the model from the stringent requirement of predicting the center.
Instead, it is sufficient for the model to predict \textit{any} salient anchor within the target moment, and the distances from the anchor to the onset and offset are predicted subsequently.
By directly locating the boundaries based on the anchor point, the model can achieve improved boundary alignment, even when the anchor point does not coincide with the actual center.

Building upon the proposed moment modeling, we introduce a new framework equipped with a dedicated decoder design, dubbed Boundary-Aligned Moment Detection Transformer (BAM-DETR).
The design of its decoding layers originates from the intuition that the refining processes for an anchor and boundaries should be distinct from each other.
That is to say, a model needs to scan over the whole video to find a potential anchor point that enables estimating the rough location of the target moment.
On the other hand, it is required to focus on fine-grained details in the vicinity to refine the boundaries to be aligned with those of the target moments.
From this motivation, in contrast to existing methods, our model adopts a dual-pathway decoding pipeline to predict an anchor point and boundaries in a parallel way.
To be specific, we leverage two different types of queries respectively for anchor and boundary refinement.
The former aggregates global information with standard attention, while the latter concentrates on the sparse local neighborhood of the boundaries using the proposed boundary-focused attention.
These distinct designs of two pathways allow for effective moment localization with minimal computational overhead increase.

\begin{table}[t]
\caption[Caption for LOF]{Impact of misaligned centers on QVHighlights.
The top-1 predictions are grouped based on their center errors normalized by ground-truth lengths. For each group, we present the mean IoU (\%) and the proportion (in parentheses). Only the predictions whose centers fall within the ground-truth moments are considered here.}
\centering
\scriptsize
\setlength{\tabcolsep}{0pt}
\begin{threeparttable}
\begin{tabularx}
{0.9\linewidth}
{@{\hspace{1mm}}p{2.6cm}p{1.4cm}<{\centering}p{1.4cm}<{\centering}p{1.4cm}<{\centering}p{1.4cm}<{\centering}p{1.4cm}<{\centering}p{1.4cm}<{\centering}}
\toprule
Method & [0, 0.1) & \selectfont[0.1, 0.2) & [0.2, 0.3) & [0.3, 0.4) & [0.4, 0.5) & All \\
\midrule
\multirow{2}{*}{Moment-DETR~\cite{lei2021qvhighlights}} & 83.20 & 65.00 & 54.82 & 43.77 & 34.98 & 67.84 \\
 & (46~\%) & (24~\%) & (14~\%) & (9~\%) & (7~\%) & (100~\%) \\
\multirow{2}{*}{QD-DETR~\cite{moon2023qd-detr}} & 87.79 & 67.95 & 55.40 & 44.93 & 36.45 & 74.09 \\
 & (56~\%) & (18~\%) & (11~\%) & (9~\%) & (5~\%) & (100~\%) \\
BAM-DETR\tnote{\dag} & 77.62 & 78.20 & 77.52 & 78.96 & 71.08 & 77.21 \\
(Ours) & (24~\%) & (21~\%) & (21~\%) & (22~\%) & (12~\%) & (100~\%)\\
\bottomrule
\end{tabularx}
\tnote{\dag}anchor points are utilized for grouping.
\end{threeparttable}
\label{tab:center_misalignment}
\end{table}


In addition, we identify the problem of the conventional scoring method, where binary classification (or matching) scores are used for proposal ranking.
This leads to suboptimal results for the grounding task since a fractional moment may have high matching scores with the given sentence.
To handle this issue, we propose to rank proposals based on their localization qualities.
Accordingly, we modify the typical matching function and training objectives of the query-based model to be localization-oriented by discarding the role of classification scores.
In this way, our model can prioritize the moment proposals exhibiting high overlap with ground truths at inference, leading to improved grounding performance.

The advantage of our BAM-DETR is showcased in \Cref{tab:center_misalignment}.
It can be observed that the anchor points predicted by our model are evenly distributed within the ground-truth intervals.
Importantly, our model consistently produces precise moments with high IoUs ($> 0.7$) across different groups, indicating that it does not depend on accurate center prediction.
On average, our model shows superior grounding performance over the existing methods with the help of our moment formulation and quality-based scoring.
In a later section, we validate the efficacy of the proposed methods through extensive experiments.
Notably, our model outperforms previous methods by large margins on three public benchmarks.



\section{Related Works}
\label{sec:related_works}

\subsection{Temporal Sentence Grounding in Videos}
Temporal sentence grounding requires seeking temporal spans semantically relevant to the given sentence in a video.
Proposal-based approaches adopt the two-stage pipeline, \ie, proposal generation and ranking.
They generate moment proposals by relying on sliding windows~\cite{gao2017tall,liu2018cross-modal,zhang2019exploiting-relationship,ge2019mac} or utilizing pre-defined anchors~\cite{zhang2019man,xu2019multilevel-integration,yuan2019semantic-modulation,chen2018temporally-grounding,zhang2019cross-interaction}.
Several works process all possible candidates at once with 2D maps~\cite{zhang20202d-tan,wang2022negative-matters,liu2021biaffine,xiao2021boundary-proposal,li2023g2l}.
Meanwhile, proposal-free methods are developed for efficient grounding by directly predicting the moments~\cite{yuan2019find-where,mun2020lgi} or estimating the probabilities of each frame being starting and ending positions~\cite{ghosh2019excl,zhang2020vslnet}.
Some approaches perform dense regression by predicting the boundaries from individual frames~\cite{zeng2020dense-regression,chen2020rethinking-bottomup,lu2019debug}.
Recently, query-based models streamline the complicated sentence grounding pipeline by removing handcrafted techniques~\cite{cao2021pursuit,liu2022umt,xu2023mh-detr,jang2023eatr,li2023momentdiff}.
There are also attempts to unify temporal sentence grounding with other video understanding tasks into a single framework~\cite{lin2023univtg,yan2023unloc}.

Our method belongs to the query-based group~\cite{lei2021qvhighlights,moon2023qd-detr}, inheriting the benefit of architectural simplicity.
In contrast to others, we employ a boundary-oriented formulation of moments to relieve the heavy reliance on center predictions, leading to better boundary alignment.
Our method also relates to dense regression methods~\cite{zeng2020dense-regression,lin2023univtg,chen2020rethinking-bottomup} that predict boundaries from each frame as an anchor.
Comparatively, our model leverages dynamic anchors that are gradually adjusted through decoding, enabling precise grounding using a small set of predictions.

\subsection{Detection Transformers}
Query-based temporal sentence grounding models by design are closely related to the family of detection transformers (DETR)~\cite{carion2020detr}.
Since the advent of DETR, a number of variants have been introduced to improve it from various perspectives~\cite{liu2021wb-detr,sun2021tsp,lin2023plain-detr,liu2023stable-dino}.
Some works focus on reducing the excessive computational costs of vanilla transformers in order to leverage multi-scale features~\cite{dai2021dynamic-detr,zhu2021deformable-detr,roh2022sparse-detr,li2023lite-detr,zheng2023focus-detr}.
On the other hand, several methods attempt to speed up the model convergence by manipulating the attention operations~\cite{gao2021modulated-coattention,liu2022dab-detr,zhang2022accelerating-convergence,ye2023cascade-detr} or incorporating the denoising process during the model training~\cite{li2022dn-detr,zhang2023dino}.

The most relevant works to ours are those which propose explicit anchor modeling for object queries using center points~\cite{meng2021conditional-detr,wang2022anchor-detr} or boxes (center, width, and height)~\cite{zhu2021deformable-detr,liu2022dab-detr}.
In comparison to an object, a moment in temporal sentence grounding has its own challenges such as center ambiguity and indistinct boundaries.
To accommodate the discrepancies, we propose a novel boundary-oriented modeling of moments to replace the conventional center-based 1D box modeling.
The advantages of our approach are clearly verified in the experiments.

\begin{figure*}[t]
    \centering
    \includegraphics[width=0.99\linewidth]{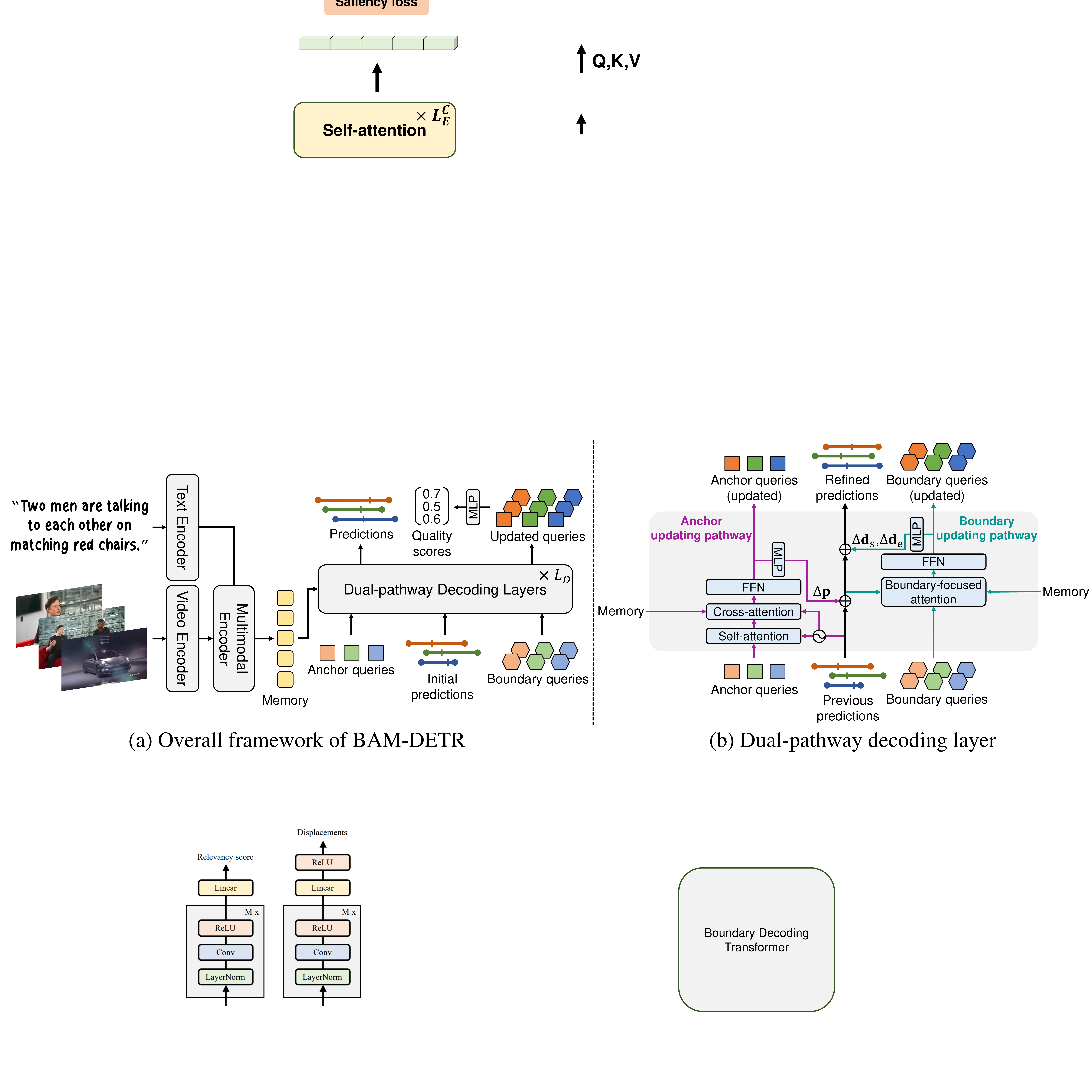}
    \caption{(a) Overview of the proposed BAM-DETR. (b) Details of the proposed dual-pathway decoding layer. It consists of two parallel pathways respectively for anchor and boundary updates, which refine previous moment predictions in a sequential manner.}
    \label{fig:architecture}
\end{figure*}

\section{Method}
\label{sec:method}
Given an untrimmed video and a sentence, the goal of temporal sentence grounding is to localize relevant moments $\{\varphi_n=(t_{s_n}, t_{e_n})\}_{n=1}^{N}$, where $N$ denotes the number of ground truths in the video and $\varphi_n$ indicates the temporal interval of the $n$-th moment.
Note that a video may have multiple moments that match the sentence, \ie, $N \geq 1$.
During the test time, the model is expected to produce a total of $M$ predictions, $\{(\hat{\varphi}_m, q_m)\}_{m=1}^{M}$, where $\hat{\varphi}_m = (\hat{t}_{s_m}, \hat{t}_{e_m})$ is the $m$-th prediction, while $q_m$ is its score for ranking.

\paragraph{Motivation.}
We tackle two main problems in the current prediction process.
On one hand, existing query-based approaches adopt the symmetric design of $(c, l)$ to predict a moment, \ie, $\hat{\varphi} = (c - 0.5l, c+0.5l)$.
As we discuss in \cref{sec:intro}, this strategy suffers from the issue of over-reliance on the center prediction, leading to unstable performance.
To address this, we propose a boundary-oriented modeling with a triplet of ($p, d_s, d_e$), where a moment is represented by $\hat{\varphi} = (p - d_s, p + d_e)$.
This asymmetric design enables direct alignment of boundaries without relying on precise center prediction.
On the other hand, previous models use classification scores as $q_m$.
This scoring is prone to sub-optimal solutions since only a fraction of the moment may well match the sentence, leading incomplete predictions to be highly ranked.
To handle this, we propose quality-based scoring to sort the proposals based on their localization qualities rather than the degree of matching.

\subsection{Overview}
As shown in \cref{fig:architecture}\textcolor{red}{a}, our BAM-DETR follows the encoder-decoder pipeline.
Briefly, it first extracts multimodal video features using the encoders.
Taking them as memory, the dual-pathway decoder predicts temporal spans and their quality scores by progressively refining learnable initial spans.

\subsection{Feature Extraction}
\label{subsec:encoder}
Following the convention~\cite{zhang20202d-tan,mun2020lgi,zhang2020vslnet}, we employ pre-trained encoders for token-level unimodal feature extraction.
It is worth noting that all unimodal encoders are kept frozen during training to avoid memory exploding.
In this stage, we obtain $D_v$-dim video features $E_v \in \R^{N_v \times D_v}$ and $D_t$-dim text features $E_t \in \R^{N_t \times D_t}$, where $N_v$ and $N_t$ are the numbers of clips and words, respectively.

\subsection{Multimodal Encoder}
We feed the unimodal features of the video and text to the multimodal encoder, so as to fuse them into text-aware video representations for temporal sentence grounding.
While various multimodal encoders are explored~\cite{lei2021qvhighlights,liu2022umt,lin2023univtg}, we adopt the text-to-video encoder design~\cite{moon2023qd-detr} consisting of cross- and self-attention blocks~\cite{vaswani2017transformer}.
Before getting into the encoder, unimodal features are projected into a shared space to facilitate cross-modal interaction, \ie, $\mathcal{V}=f_v(E_v) \in \R^{N_v \times D}$, $\mathcal{T}=f_t(E_t) \in \R^{N_t \times D}$, where $D$ is the embedding dimension.
Afterward, multi-head cross-attention blocks are used to inject textual information into the clip-level video representations.
In specific, we project $\mathcal{V}$ to $\mathbf{Q}_{\mathcal{V}}$ (query) while $\mathcal{T}$ to $\mathbf{K}_{\mathcal{T}}$ (key) and $\mathbf{V}_{\mathcal{T}}$ (value).
Then a cross-attention block can be formulated as:
\begingroup
\setlength{\abovedisplayskip}{3pt}
\setlength{\belowdisplayskip}{3pt}
\begin{equation}
\begin{aligned}
    \mathcal{V}' &= \text{softmax}\Big(\frac{\mathbf{Q}_{\mathcal{V}}\mathbf{K}_{\mathcal{T}}^\top}{\sqrt{D}}\Big)\mathbf{V}_{\mathcal{T}}  + \mathcal{V}, \\
    \mathcal{V}'' &= \text{FFN}(\mathcal{V'}) + \mathcal{V}',
\end{aligned}
\label{eq:attn_block}
\end{equation}
\endgroup
where $\text{FFN}(\cdot)$ is a feed-forward network.
Although we here present the single-head attention block, it can readily generalize to a multi-headed version~\cite{vaswani2017transformer}.
We denote the resulting multimodal representations obtained after $L_E$ multi-head cross-attention blocks by $\tilde{\mathcal{V}} \in \R^{N_v \times D}$.

Subsequently, self-attention blocks are leveraged to enhance the representations by allowing the inter-clip interaction.
Here we project $\tilde{\mathcal{V}}$ to $\mathbf{Q}_{\tilde{\mathcal{V}}}$, $\mathbf{K}_{\tilde{\mathcal{V}}}$, and $\mathbf{V}_{\tilde{\mathcal{V}}}$.
We note that the query and the key are supplemented with fixed sinusoidal positional encoding~\cite{vaswani2017transformer,carion2020detr} for temporal awareness.
Then the self-attention block is defined in a similar way to \cref{eq:attn_block} but with different inputs.
The enhanced clip-level representations after $L_{E}$ multi-head self-attention blocks are denoted by $\hat{\mathcal{V}} \in \R^{N_v \times D}$, which will serve as the \textit{memory} for the decoder.

It is widely known that providing saliency guidance to the memory features helps the model to better understand the semantic relationship between the video and text~\cite{lei2021qvhighlights,lin2023univtg}.
As in previous works~\cite{jang2023eatr,li2023momentdiff}, we impose saliency score constraints on the memory features.
Specifically, we leverage a saliency predictor $S(\cdot)$ and train the model with the following margin-based training objective.
\begin{equation}
    \mathcal{L}_{\text{margin}} = \max(0, \alpha + S(\hat{v}^{\text{low}}) - S(\hat{v}^{\text{high}})),
\label{eq:loss_margin}
\end{equation}
where $\alpha$ is a margin and $(\hat{v}^{\text{low}}, \hat{v}^{\text{high}})$ is the sampled feature pair satisfying that the saliency label of $\hat{v}^{\text{low}}$ is lower than that of $\hat{v}^{\text{high}}$.
In case of the absence of saliency labels, we collect clips within and outside the ground-truth moment intervals to build a pair.
In addition, we employ the rank-aware contrastive loss $\mathcal{L}_{\text{cont}}$ and the negative relation loss $\mathcal{L}_{\text{neg}}$, following Moon~\etal~\cite{moon2023qd-detr}.
Due to space limits, we refer the readers to Appendix for the loss formulations.
In summary, the overall saliency loss is defined as $\mathcal{L}_{\text{sal}} = \mathcal{L}_{\text{margin}} + \mathcal{L}_{\text{cont}} + \mathcal{L}_{\text{neg}}$.

\subsection{Dual-pathway Decoder}
\label{subsec:decoder}

With the multimodal representations $\hat{\mathcal{V}}$ as memory features, we aim to localize temporal spans corresponding to the sentence.
We adopt the new boundary-oriented formulation for moment prediction, where each prediction is represented by a triplet of $(p, d_s, d_e)$, where $p$ is the anchor point, while $d_s$ and $d_e$ are the distances from the anchor to the starting and ending points, respectively.
To make full use of the proposed formulation, we design a dual-pathway decoding layer with two parallel pathways (\cref{fig:architecture}\textcolor{red}{b}).
Formally, the inputs of the $l$-th layer are anchor queries $\mathbf{C}_p^l \in \R^{M \times D}$, boundary queries $\mathbf{C}_s^l, \mathbf{C}_e^l \in \R^{M \times D}$, and previous moment predictions $\mathbf{A}^l = \begin{bmatrix}\mathbf{p}^l; \mathbf{d}_s^l; \mathbf{d}_e^l\end{bmatrix} \in \R^{M \times 3}$, where $M$ is the number of queries (predictions).
Note that the initial queries and spans, \ie, $\{\mathbf{C}_p^{0}$, $\mathbf{C}_s^{0}$, $\mathbf{C}_e^{0}$, $\mathbf{A}^{0}\}$, are learnable parameters.
We elaborate on the two pathways in the following.

\paragraph{Anchor updating pathway.}
Given the predictions and the anchor queries from the preceding layer, the goal of this pathway is to adjust the anchor position so that the boundaries can be predicted based on it.
Intuitively, in order to obtain a valuable anchor point without redundancy, an anchor query should communicate with other queries as well as the memory features.
To this goal, the anchor updating pathway consists of a self-attention layer, a cross-attention layer, and a feed-forward network.
In the self-attention layer, the anchor queries $\mathbf{C}_p^l$ are first projected into $\mathbf{Q}_{\mathbf{C}_p^l}$, $\mathbf{K}_{\mathbf{C}_p^l}$, and $\mathbf{V}_{\mathbf{C}_p^l}$.
Since an anchor query itself lacks positional information, we build positional encoding.
Following the previous works~\cite{moon2023qd-detr,jang2023eatr}, we extend the current spans $\mathbf{A}^l$ to the positional information of the queries, \ie, $\mathbf{P}_{\mathbf{A}^l}=\text{MLP}(\text{PE}(\mathbf{A}^l)) \in \R^{M \times D}$, where $\text{PE}(\cdot)$ denotes the point-wise mapping from a position to the corresponding sinusoidal encoding while $\text{MLP}(\cdot)$ is multi-layer perceptron.
The self-attention for anchor queries is defined as:
\begingroup
\setlength{\abovedisplayskip}{5pt}
\setlength{\belowdisplayskip}{5pt}
\begin{equation}
    \tilde{\mathbf{C}}_p^l = \text{softmax}\Big(\frac{\big(\mathbf{Q}_{\mathbf{C}_p^l} + \mathbf{P}_{\mathbf{A}^l}\big)\big(\mathbf{K}_{\mathbf{C}_p^l} + \mathbf{P}_{\mathbf{A}^l}\big)^\top}{\sqrt{D}}\Big)\mathbf{V}_{\mathbf{C}_p^{l}}  + \mathbf{C}_p^l.
\label{eq:attn_block_2}
\end{equation}
\endgroup
After inter-query interaction, we employ a global cross-attention layer to aggregate multi-modal features from the memory.
The anchor queries $\tilde{\mathbf{C}}_p^l$ is projected into $\mathbf{Q}_{\tilde{\mathbf{C}}_p^l}$ while the memory $\hat{\mathcal{V}}$ is projected to $\mathbf{K}_{\hat{\mathcal{V}}}$ and $\mathbf{V}_{\hat{\mathcal{V}}}$.
To make the query location-aware, we leverage the sinusoidal encoding of current anchor positions, \ie, $\mathbf{P}_{\mathbf{p}^l}=\text{PE}(\mathbf{p}^l) \in \R^{M \times D}$.
Similarly, the memory leverages the positional encoding, \ie, $\mathbf{P}_{\hat{\mathcal{V}}}=\text{PE}(\hat{\mathcal{V}}) \in \R^{N_v \times D}$.
We use concatenation instead of summation to separate the roles of features and positional encoding~\cite{meng2021conditional-detr,liu2022dab-detr}.
The cross-attention between anchor queries and the memory can be expressed as:
\begingroup
\setlength{\abovedisplayskip}{5pt}
\setlength{\belowdisplayskip}{5pt}
\begin{equation}
    \hat{\mathbf{C}}_p^l = \text{softmax}\Big(\frac{\big(\mathbf{Q}_{\tilde{\mathbf{C}}_p^l} \mathbin\Vert \mathbf{P}_{\mathbf{p}^l}\big)\big(\mathbf{K}_{\hat{\mathcal{V}}} \mathbin\Vert \mathbf{P}_{\hat{\mathcal{V}}})\big)^\top}{\sqrt{2D}}\Big)\mathbf{V}_{\hat{\mathcal{V}}}  + \tilde{\mathbf{C}}_p^l.
\label{eq:attn_block_3}
\end{equation}
\endgroup
After all, the anchor queries are updated with a feed-forward network, \ie, $\mathbf{C}_p^{(l+1)} = \text{FFN}(\hat{\mathbf{C}}_p^l) + \hat{\mathbf{C}}_p^l$.
Lastly, we adjust the anchor positions using sigmoid-based refinement~\cite{zhu2021deformable-detr}: $\hat{\mathbf{A}}^l = \begin{bmatrix}\mathbf{p}^{(l+1)}; \mathbf{d}_s^l; \mathbf{d}_e^l \end{bmatrix}$ where $\mathbf{p}^{(l+1)} = \sigma(\sigma^{-1}(\mathbf{p}^l) + \Delta \mathbf{p}^l)$ with $\Delta \mathbf{p}^l = \text{MLP}(\mathbf{C}_p^{(l+1)}) \in \R^{M}$ and the sigmoid function $\sigma(\cdot)$.

\begin{figure*}[t]
    \centering
    \includegraphics[width=0.62\linewidth]{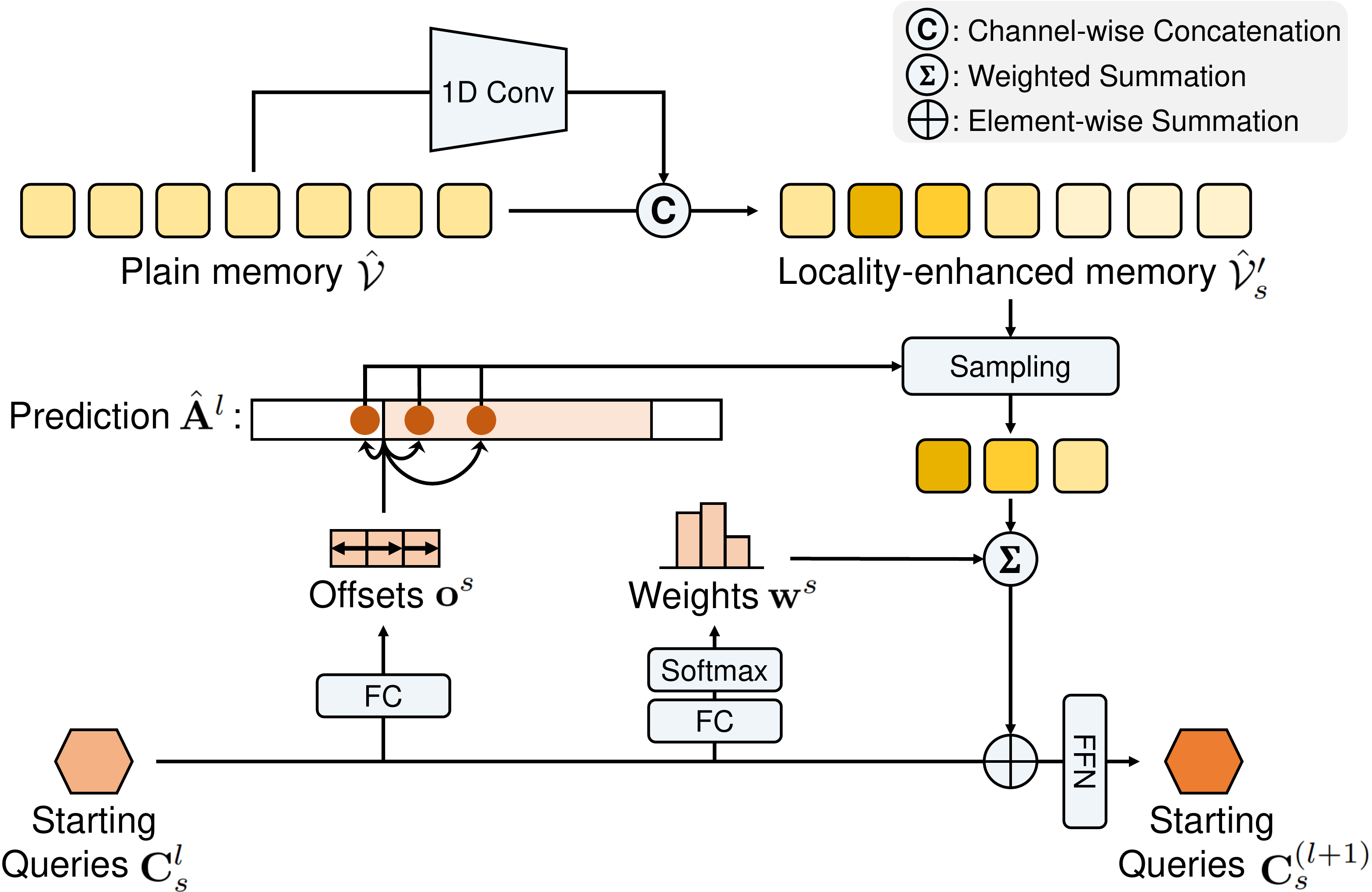}
    \caption{Boundary-focused attention layer for starting queries.}
    \label{fig:boundary_layer}
\end{figure*}

\paragraph{Boundary updating pathway.}
After the anchor update, we refine the boundaries of the predictions.
It is widely perceived that a model needs to focus on fine-grained features in the neighborhood rather than far ones to adjust temporal boundaries~\cite{lin2018bsn,lin2021afsd,shi2023tridet}.
Inspired by this, we devise a boundary-focused attention layer (\cref{fig:boundary_layer}).
For brevity, we explain the process of starting boundary update.

The plain memory $\hat{\mathcal{V}}$ lacks the inductive bias due to the property of attentional layers~\cite{dosovitskiy2021vit}.
Thus we first build locality-enhanced memory features for effective boundary refinement.
To this goal, we obtain boundary-sensitive features with several 1D convolutional layers, \ie, $\hat{\mathcal{V}}_{\text{s}}=f_{\text{s}}(\hat{\mathcal{V}})$.
Then we encourage them to highly activate around the starting position of target moments.
In detail, we impose regularization on the clip-wise activation scores obtained by channel-mean, \ie, $\hat{g}^s = \text{mean}(\sigma(\hat{\mathcal{V}_s})) \in \R^{N_v}$.
The regularization loss is defined as:
\begingroup
\setlength{\abovedisplayskip}{3pt}
\setlength{\belowdisplayskip}{3pt}
\begin{equation}
    \mathcal{L}_{\text{regul}}^s = -\frac{1}{N_v}\sum_{i=1}^{N_v}\big(g_i^s \text{log}(\hat{g}_i^s) + (1 - g_i^s) \text{log}(1 - \hat{g}_i^s)\big),
\label{eq:loss_regul}
\end{equation}
\endgroup
where $g_i^s$ is the binary label obtained as $g_i^s = \mathbbm{1} [i \in \mathcal{B}^s]$, where $\mathcal{B}^s$ is the neighbor clip set around starting points with a radius of $r^s$ (set to $1/10$ of the moment length).
We merge boundary-sensitive features with the plain ones to form locality-enhanced memory, \ie, $\hat{\mathcal{V}}'_s = [\hat{\mathcal{V}} \mathbin\Vert \hat{\mathcal{V}}_s ] \in \R^{N_v \times 2D}$.
We can obtain $\mathcal{L}_{\text{regul}}^e$ and $\hat{\mathcal{V}}'_e$ in the same way.
The total regularization term is $\mathcal{L}_\text{regul} = \mathcal{L}_\text{regul}^s + \mathcal{L}_\text{regul}^e$.

Given the locality-enhanced memory, the boundary queries need to capture fine-grained details around the boundaries for refinement.
To efficiently aggregate useful features near the boundaries, we employ deformable attention~\cite{zhu2021deformable-detr,zhu2019deformable-conv2}.
Regarding the starting boundary, \ie, $\mathbf{p}^{(l+1)}-\mathbf{d}_s^l$, as the origin, we predict offsets and weights to select $K$ neighbors, \ie, $\mathbf{o}^s = \phi_o(\mathbf{C}_{s}^{l}) \in \R^{M \times K}$, $\mathbf{w}^s = \text{softmax}(\phi_w(\mathbf{C}_{s}^{l})) \in \R^{M \times K}$, where $\phi_*$ are fully-connected layers.
Features from the sampled neighbors are then aggregated into the starting queries as:
\begingroup
\setlength{\abovedisplayskip}{3pt}
\setlength{\belowdisplayskip}{3pt}
\begin{equation}
    \hat{\mathbf{C}}_s^l = \sum_{k=1}^{K}\big[\mathbf{w}_k^s \cdot \hat{\mathcal{V}}'_s[\mathbf{p}^{(l+1)}-\mathbf{d}_s^l + \mathbf{o}_k^s]\big] + \mathbf{C}_s^l,
\label{eq:deform_attn}
\end{equation}
\endgroup
where we denote the sampling process from memory by $\hat{\mathcal{V}}'_s[\cdot]$.
Lastly, we adopt a feed-forward network to obtain the updated starting queries, \ie, $\mathbf{C}_s^{(l+1)}=\text{FFN}(\hat{\mathbf{C}}_s^l)$.
We also obtain $\mathbf{C}_e^{(l+1)}$ in the same way.
With the updated queries, we refine the boundaries to be better aligned with those of ground truths using sigmoid-based refinement as similar in the anchor update, leading to the refined predictions $\mathbf{A}^{(l+1)} = [\mathbf{p}^{(l+1)}; \mathbf{d}_s^{(l+1)}; \mathbf{d}_e^{(l+1)} ]$.
Note that we utilize the deformable attention for the purpose of local feature aggregation, which sharply differs from the original purpose of efficient multi-scale global operation~\cite{zhu2021deformable-detr}.
We provide comparison experiments regarding their roles in Appendix.

\paragraph{Moment prediction.}
We repeatedly update the predictions through a total of $L_D$ dual-pathway decoding layers.
We denote the resulting predictions by $\mathbf{A}=\begin{bmatrix}\mathbf{p}; \mathbf{d}_s; \mathbf{d}_e\end{bmatrix}$, the anchor queries by $\mathbf{C}_p$, and the boundary queries by $\mathbf{C}_s$ and $\mathbf{C}_e$.
Then we cast the predictions in the form of starting and ending timestamps, \ie, $\hat{\varphi} = \begin{bmatrix}\mathbf{p}-\mathbf{d}_s; \mathbf{p}+\mathbf{d}_e\end{bmatrix} \in \R^{M \times 2}$, which serves as the final results.

\subsection{Quality-based Scoring}
\label{subsec:quality_scoring}
After producing the moment predictions, we opt to rank them for evaluation.
In the convention, query-based models utilize classification scores as the measure, which exhibits how well the proposals semantically match the sentence.
However, it does not necessarily represent the localization qualities of proposals.
Hence we propose to estimate the localization quality of each moment prediction.
Formally, the quality score can be derived as $\mathbf{q} = \sigma(\text{MLP}([\mathbf{C}_p \mathbin\Vert \mathbf{C}_s \mathbin\Vert \mathbf{C}_e])) \in \R^{M}$, where $\sigma$ is sigmoid activation.
Then the quality loss is defined as follows.
\begin{equation}
    \mathcal{L}_{\text{qual}} = \sum_{m=1}^{M} \Big| q_m - \max_{\forall n}\big(\frac{|\hat{\varphi}_m \cap \varphi_n|}{|\hat{\varphi}_m \cup \varphi_n|}\big)\Big|, 
\end{equation}
\noindent where the objective of the quality head is to predict the maximum IoUs of the proposals with ground-truth moments.

\subsection{Matching}
\label{subsec:training_inference}
As in the standard of query-based models~\cite{carion2020detr,lei2021qvhighlights}, we perform Hungarian matching~\cite{kuhn1955hungarian} between predictions and ground truths.
The optimal matching results $\psi^*$ can be derived as follows.
\begingroup
\setlength{\abovedisplayskip}{3pt}
\setlength{\belowdisplayskip}{3pt}
\begin{equation}
\begin{aligned}
    &\psi^* = {\arg\min}_{\psi \in \mathfrak{G}_N} \sum_{n=1}^{N} \mathcal{C}(\varphi_{n}, \hat{\varphi}_{\psi(n)}), \\
    \mathcal{C}(\varphi_{n}, \hat{\varphi}_{\psi(n)}) &= \lambda_{l_1}\mathcal{L}_{l_1}(\varphi_{n}, \hat{\varphi}_{\psi(n)}) + \lambda_{\text{iou}}\mathcal{L}_{\text{iou}}(\varphi_{n}, \hat{\varphi}_{\psi(n)}),
\end{aligned}
\label{eq:cost_func}
\end{equation}
\endgroup
where $\mathfrak{G}_N$ denotes the combination pool and $\psi(n)$ is the index of the prediction matched by the $n$-th ground truth.
$\mathcal{L}_{l_1}$ and $\mathcal{L}_{\text{iou}}$ respectively represent the $L_1$ distance and the generalized IoU~\cite{rezatofighi2019giou} between the moments, while $\lambda_{l_1}$ and $\lambda_{\text{iou}}$ are their weights.
Note that in contrast to other works, the classification term is not involved in the matching process, leading to localization-oriented matching.

Once the matching is completed, we minimize the matching cost between each pair of the matching results $\psi^*$.
The localization loss is defined as:
\begin{equation}
    \mathcal{L}_{\text{loc}} = \sum_{n=1}^{N} \mathcal{C}(\varphi_n, \hat{\varphi}_{\psi^*(n)}).
\end{equation}

\paragraph{Overall training objectives.}
Our model is trained in an end-to-end fashion and the overall training objective is defined as follows.
\begingroup
\setlength{\abovedisplayskip}{3pt}
\setlength{\belowdisplayskip}{3pt}
\begin{equation}
    \mathcal{L}_{\text{total}} = \mathcal{L}_{\text{loc}} + \lambda_{\text{qual}} \mathcal{L}_{\text{qual}} + \lambda_{\text{sal}} \mathcal{L}_{\text{sal}} + \lambda_{\text{regul}} \mathcal{L}_{\text{regul}},
\end{equation}
\endgroup
where $\lambda_{*}$ are the balancing parameters.
\section{Experiments}
\label{sec:experiments}

\subsection{Experimental Settings}
\noindent\textbf{Datasets.}~
QVHighlights~\cite{lei2021qvhighlights} is a recently built dataset, containing a total of 10,148 videos and 10,310 sentences from vlog and news domains.
In addition to moments, it provides segment-level saliency score annotations within each moment.
Importantly, it allows a single sentence corresponding to multiple disjoint moments (1.8 on average).
Due to this practical setup, we utilize QVHighlights as the main benchmark.
Charades-STA~\cite{gao2017tall} includes 16,128 sentence-moment pairs with 9,848 indoor videos.
The average duration of videos and moments is 30.6 and 8.1 seconds, respectively.
TACoS~\cite{regneri2013tacos} contains 127 cooking videos encompassing a total of 18,818 sentence-moment pairs.
This dataset is known to be challenging since the moments occupy only a small portion (6.1 sec on average) within considerably long videos (4.8 min on average).

\noindent\textbf{Evaluation metrics.}~
Following the standard protocol, we measure the Recall@1 (R1) under the IoU thresholds of 0.3, 0.5, and 0.7 by default.
Since QVHighlights contains multiple ground-truth moments per sentence, we report the mean average precision (mAP) with IoU thresholds of 0.5 and 0.75 as well as the average mAP over a set of IoU thresholds [0.5:0.05:0.95].
Meanwhile, we compute the mean IoU of top-1 predictions on Charades-STA and TACoS.
Note that the performances at high IoU thresholds (\eg, 0.7) exhibit how well the predictions align with the ground truths.

\begin{table}[t]
\centering
\scriptsize
\setlength{\tabcolsep}{0pt}
\begin{threeparttable}
\begin{tabularx}{0.77\linewidth}{@{\hspace{2mm}}p{3.5cm}p{1.1cm}<{\centering}p{1.1cm}<{\centering}p{1.0mm}<{\centering}p{1.1cm}<{\centering}p{1.1cm}<{\centering}p{1.1cm}<{\centering}p{1.0mm}}
\toprule
\multirow{2}{*}{Method}
& \multicolumn{2}{c}{R1} & & \multicolumn{3}{c}{mAP} \\
\cmidrule{2-3} \cmidrule{5-7}
& @0.5 & @0.7 & & @0.5 & @0.75 & Avg. \\
\midrule
\multicolumn{7}{l}{\hspace{0.5mm}\textit{w/o pre-training}} \\
MCN~\cite{anne2017mcn} & 11.41 & 2.72 & & 24.94 & 8.22 & 10.67 \\
CAL~\cite{escorcia2019cal} & 25.49 & 11.54 & & 23.40 & 7.65 & 9.89 \\
CLIP~\cite{radford2021clip} & 16.88 & 5.19 & & 18.11 & 7.00 & 7.67 \\
XML~\cite{lei2020tvr}   & 46.69 & 33.46 & & 47.89 & 34.67 & 34.90 \\
Moment-DETR~\cite{lei2021qvhighlights} & 52.89 & 33.02 & & 54.82 & 29.40 & 30.73 \\
UMT\tnote{\dag}~\cite{liu2022umt}  & 56.23 & 41.18 & & 53.83 & 37.01 & 36.12 \\
MH-DETR~\cite{xu2023mh-detr}  & 60.05 & 42.48 & & 60.75 & 38.13 & 38.38 \\
QD-DETR~\cite{moon2023qd-detr}  & 62.40 & 44.98 & & 62.52 & 39.88 & 39.86 \\
QD-DETR\tnote{\dag}~\cite{moon2023qd-detr}  & 63.06 & 45.10 & & 63.04 & 40.10 & 40.19 \\
UniVTG~\cite{lin2023univtg} & 58.86 & 40.86 & & 57.60 & 35.59 & 35.47 \\
EaTR\tnote{\ddag}~\cite{jang2023eatr} & 57.98 & 42.41 & & 59.95 & 39.29 & 39.00 \\
MomentDiff~\cite{li2023momentdiff} & 57.42 & 39.66 & & 54.02 & 35.73 & 35.95 \\
\cmidrule{1-7}
{\ours} & \textbf{62.71} & \textbf{48.64} & & \textbf{64.57} & \textbf{46.33} & \textbf{45.36} \\
{\ours}\tnote{\dag} & \textbf{64.07} & \textbf{48.12} & & \textbf{65.61} & \textbf{47.51} & \textbf{46.91} \\
\midrule 
\multicolumn{7}{l}{\hspace{0.5mm}\textit{w/ pre-training on 4.2M data labeled by CLIP}} \\
UniVTG~\cite{lin2023univtg} & 65.43 & 50.06 & & 64.06 & 45.02 & 43.63 \\
\midrule
\multicolumn{7}{l}{\hspace{0.5mm}\textit{w/ pre-training on 236K ASR captions}} \\
Moment-DETR~\cite{lei2021qvhighlights}  & 59.78 & 40.33 & & {60.51} & 35.36 & 36.14 \\
UMT\tnote{\dag}~\cite{liu2022umt} & {60.83} & {43.26} & & 57.33 & {39.12} & {38.08} \\
QD-DETR~\cite{moon2023qd-detr} & 63.18 & 45.19 & & 63.37 & 40.35 & 39.96 \\
\cmidrule{1-7}
{\ours} & \textbf{63.88} & \textbf{47.92} & & \textbf{66.33} & \textbf{48.22} & \textbf{46.67} \\
\bottomrule
\end{tabularx}
\tnote{\dag}additional use of audio modality~~\tnote{\ddag}reproduced by official checkpoint
\end{threeparttable}
\caption{Results on the QVHighlights test split.}
\label{tab:qvhighlights}
\end{table}

\subsection{Implementation Details}
For a fair comparison, we adopt the same feature extraction strategy with the competitors~\cite{lei2021qvhighlights,moon2023qd-detr,li2023momentdiff}.
Specifically, we adopt the CLIP~\cite{radford2021clip} text features for text and the concatenation of Slowfast~\cite{feichtenhofer2019slowfast} (ResNet-50) and CLIP~\cite{radford2021clip} (ViT-B/32) features for videos unless otherwise specified.
The video features are extracted every 1 second for Charades-STA and 2 seconds for QVHighlihgts and TACoS.
To compare with audio-augmented models~\cite{liu2022umt}, we optionally employ audio features extracted by PANNs~\cite{kong2020panns} pre-trained on AudioSet~\cite{gemmeke2017audioset}.
Due to the long video duration, we uniformly sample 200 feature vectors from each video for TACoS.

We set the embedding dimension $D$ to 256, the number of attention heads to 8, the number of queries $M$ to 10, the number of boundary points $K$ to 3, and the margin $\alpha$ to 0.2.
We determine the numbers of encoding and decoding layers as same with the prior work~\cite{moon2023qd-detr}, \ie, $L_E = L_D = 2$.
The balancing parameters are set as: $\lambda_{l1} = 10$, $\lambda_{\text{iou}} = \lambda_{\text{sal}} = \lambda_{\text{regul}} = 1$, $\lambda_{\text{qual}} = 2$.
As in previous works~\cite{li2023momentdiff,moon2023qd-detr}, we increase $\lambda_{\text{sal}}$ to 4 when saliency labels are unavailable, \ie, for Charades-STA and TACoS.
Our model is trained from scratch for 200 epochs on QVHighlights and 100 epochs on the other datasets using the AdamW optimizer~\cite{loshchilov2019adamw} with a learning rate of 1e-4 and a batch size of 32.

\subsection{Comparison with State-of-the-arts}
\noindent\textbf{Results on QVHighlights.}~
We compare our model with existing state-of-the-arts including recent query-based approaches~\cite{lei2021qvhighlights,liu2022umt,xu2023mh-detr,moon2023qd-detr,jang2023eatr,li2023momentdiff} on the test split.
As shown in \Cref{tab:qvhighlights}, our BAM-DETR consistently outperforms the comparative models under various settings.
In detail, without pre-training, our model surpasses the previous state-of-the-art model~\cite{moon2023qd-detr} by large margins, \eg, 3.66\% in R1@0.7 and 6.45\% in mAP@0.75.
These improvements under strict IoU thresholds verify the superior localization ability of our method.
When leveraging auxiliary audio features, the performance further boosts especially in R1@0.5 and mAPs, enlarging the gap between the competitors including those with audio features~\cite{liu2022umt,moon2023qd-detr}.
To compare with the methods with pretraining, we pretrain our model on middle-scale ASR caption data~\cite{lei2021qvhighlights}.
Again, our BAM-DETR achieves state-of-the-art results, while showing the least gap with the method~\cite{lin2023univtg} that leverages large-scale data for pre-training.
Notably, our model even outperforms it in terms of mAPs with much fewer (about 18$\times$) pre-training data, manifesting the effectiveness of the proposed methods.

\begin{table}
\centering
\scriptsize
\setlength{\tabcolsep}{0pt}
\begin{tabularx}
{0.98\linewidth}
{@{\hspace{2mm}}p{2.5cm}p{1.15cm}<{\centering}p{1.15cm}<{\centering}p{1.15cm}<{\centering}p{1.05cm}<{\centering}p{1.0mm}<{\centering}p{1.15cm}<{\centering}p{1.15cm}<{\centering}p{1.15cm}<{\centering}p{1.15cm}<{\centering}p{1.05cm}<{\centering}}
\toprule

\multirow{2}{*}{Method}& \multicolumn{4}{c}{Charades-STA} & & \multicolumn{4}{c}{TACoS} \\
\cmidrule{2-5} \cmidrule{7-10}
& R1@0.3 & R1@0.5 & R1@0.7 & mIoU & & R1@0.3 & R1@0.5 & R1@0.7 & mIoU \\
\midrule
2D-TAN~\cite{zhang20202d-tan} & 58.76 & 46.02 & 27.40 & 41.25 & & 40.01 & 27.99 & 12.92 & 27.22 \\
VSLNet~\cite{zhang2020vslnet} & 60.30 & 42.69 & 24.14 & 41.58 & & 35.54 & 23.54 & 13.15 & 24.99 \\
Moment-DETR~\cite{lei2021qvhighlights} & 65.83 & 52.07 & 30.59 & 45.54 & & 37.97 & 24.67 & 11.97 & 25.49 \\
QD-DETR~\cite{moon2023qd-detr} & - & 57.31 & 32.55 & - & & 52.39 & 36.77 & 21.07 & 35.76 \\
UniVTG~\cite{lin2023univtg} & 70.81 & 58.01 & 35.65 & 50.10 & & 51.44 & 34.97 & 17.35 & 33.60 \\
MomentDiff~\cite{li2023momentdiff} & - & 55.57 & 32.42 & - & & 46.64 & 28.92 & 12.37 & 30.36 \\
\midrule
{\ours} & \textbf{72.93} & \textbf{59.95} & \textbf{39.38} & \textbf{52.33} & & \textbf{56.69} & \textbf{41.54} & \textbf{26.77} & \textbf{39.31} \\
\bottomrule
\end{tabularx}
\caption{Results on the Charades-STA and TACoS test splits.}
\label{tab:charades}
\end{table}

\noindent\textbf{Results on Charades-STA.}~
Experimental results on the test split are shown in \Cref{tab:charades}.
Not only does our BAM-DETR outperform the query-based competitors~\cite{lei2021qvhighlights,moon2023qd-detr,li2023momentdiff}, but it achieves a new state-of-the-art by surpassing the best performing anchor-free model~\cite{lin2023univtg} for all metrics.
Notably, a large gap of 3.73\% is observed in R1@0.7, which confirms the strong localization ability of our model.

\noindent\textbf{Results on TACoS.}~
We present the comparison results on the test set in \Cref{tab:charades}.
It can be observed that our BAM-DETR achieves a new state-of-the-art with pronounced performances under the strict IoU thresholds, which is consistent with the above results on other datasets.
On this challenging benchmark, our model surpasses the previous best model~\cite{moon2023qd-detr} by 5.7\% (relatively 27\%) in R1@0.7.
These results clearly exhibit the superiority of the proposed model.

\subsection{Robustness Evaluation}
Query-based models potentially have a temporal bias~\cite{yuan2021closer-look,hao2022shuffling} against the locations and lengths of moments.
To measure robustness, we evaluate our model on the anti-biased Charades-STA~\cite{li2023momentdiff} with distribution shifts of the moment location and length between training and test sets.
\Cref{tab:charades_bias} summarizes the results, where our model outperforms all competitors under both anti-biased settings.
Especially, it shows significant performance gaps under the moment length bias.
This can be expected since our model directly localizes boundaries instead of predicting lengths, which lessens the effect of bias.
This robustness test corroborates the advantage of our boundary-oriented moment modeling.

\begin{table}[t]
\centering
\scriptsize
\setlength{\tabcolsep}{0pt}
\begin{threeparttable}
\begin{tabularx}
{0.98\linewidth}
{@{\hspace{2mm}}p{2.5cm}p{1.15cm}<{\centering}p{1.15cm}<{\centering}p{1.15cm}<{\centering}p{1.05cm}<{\centering}p{1.0mm}<{\centering}p{1.15cm}<{\centering}p{1.15cm}<{\centering}p{1.15cm}<{\centering}p{1.15cm}<{\centering}p{1.05cm}<{\centering}}
\toprule

\multirow{2}{*}{Method}
& \multicolumn{4}{c}{w.r.t. moment location} & & \multicolumn{4}{c}{w.r.t. moment length} \\
\cmidrule{2-5} \cmidrule{7-10}
& R1@0.3 & R1@0.5 & R1@0.7 & mAP$_{avg}$ & & R1@0.3 & R1@0.5 & R1@0.7 & mAP$_{avg}$ \\
\midrule
2D-TAN~\cite{zhang20202d-tan} & 27.81 & 20.44 & 10.84 & 17.23 & & 39.68 & 28.68 & 17.72 & 22.79 \\
MMN~\cite{wang2022negative-matters} & 33.58 & 27.20 & 14.12 & 19.18 & & 43.58 & 34.31 & 19.94 & 26.85 \\
Moment-DETR~\cite{lei2021qvhighlights} & 29.94 & 22.16 & 11.56 & 18.66 & & 42.73 & 34.39 & 16.12 & 24.02\\
QD-DETR\tnote{\ddag}~\cite{moon2023qd-detr} & 56.17 & 46.82 & 28.13 & 30.70 & & 67.39 & 54.44 & 32.87 & 36.99 \\ 
MomentDiff~\cite{li2023momentdiff} & 48.39 & 33.59 & 15.71 & 21.37 & & 51.25 & 38.32 & 23.38 & 28.19 \\ 
\midrule
{\ours} & \textbf{59.83} & \textbf{50.00} & \textbf{32.08} & \textbf{31.68} & & \textbf{68.40} & \textbf{55.46} & \textbf{40.74} & \textbf{43.21} \\
\bottomrule
\end{tabularx}
\tnote{\ddag}reproduced by official codebase
\end{threeparttable}
\caption{Results on the anti-biased Charades-STA test split against the moment location and length. VGG~\cite{simonyan2015vgg} and Glove~\cite{pennington2014glove} features are employed for all models.}
\label{tab:charades_bias}
\end{table}

\begin{table}[t]
\centering
\scriptsize
\setlength{\tabcolsep}{0pt}
\begin{tabularx}{0.79\linewidth}{@{\hspace{2mm}}p{4.3cm}p{1cm}<{\centering}p{1cm}<{\centering}p{1.0mm}<{\centering}p{1cm}<{\centering}p{1cm}<{\centering}p{1cm}<{\centering}p{1.0mm}}
\toprule

\multirow{2}{*}{Method} & \multicolumn{2}{c}{R1} & & \multicolumn{3}{c}{mAP} \\
\cmidrule{2-3} \cmidrule{5-7}
& @0.5 & @0.7 & & @0.5 & @0.75 & Avg. \\
\midrule
Baseline & 62.39 & 47.87 & & 62.64 & 41.54 & 41.75 \\
\hspace{1mm}$+$ boundary-oriented modeling & 63.42 & 49.23 & & 62.86 & 43.24 & 42.42 \\
\hspace{1mm}$+$ dual-pathway decoder & 63.61 & 50.26 & & 63.01 & 44.98 & 44.16 \\
\hspace{1mm}$+$ quality-based scoring & 65.10 & 51.61 & & 65.41 & 48.56 & 47.61 \\
\bottomrule
\end{tabularx}
\caption{Ablation study of components on QVHighlights.}
\label{tab:component_ablation}
\end{table}

\newcommand{\TableAblationNumQueries}{
    \toprule
    
     & \multicolumn{2}{c}{R1} & & mAP \\
    \cline{2-3} \cline{5-5}    
    \vspace{-0.38cm}$M$    &@0.5 &@0.7 & & Avg. \\
    \midrule
        5 & 62.19  & 49.03  & & 45.18  \\
        \rowcolor{defaultcolor}
        10 & 65.10  & 51.61  & & 47.61 &  \\
        15 & 65.48  & 50.32  & & 48.23  \\
        20 & 65.23  & 51.61  & & 48.01  \\
    \bottomrule
}

\newcommand{\TableAblationBoundaryFeat}{
    \toprule
    
      & \multicolumn{2}{c}{R1} & & mAP \\
    \cline{2-3} \cline{5-5}    
    \vspace{-0.38cm}Memory    & @0.5 & @0.7 & & Avg. \\
    \midrule
        plain & 63.74  & 49.68  & & 46.21  \\
        sensitive & 63.81  & 49.87  & & 46.34  \\
        \rowcolor{defaultcolor}
        merged & 65.10  & 51.61  & & 47.61 &  \\
    \bottomrule
}

\newcommand{\TableAblationNumBoundaryPoint}{
    \toprule
    
      & \multicolumn{2}{c}{R1} & & mAP \\
    \cline{2-3} \cline{5-5}    
    \vspace{-0.38cm}$K$    & @0.5 & @0.7 & & Avg. \\
    \midrule
        1 (fixed) & 63.23  & 48.32  & & 45.17  \\
        1 & 64.32  & 49.87  & & 47.14  \\
        \rowcolor{defaultcolor}
        3 & 65.10  & 51.61  & & 47.61 &  \\
        5 & 64.90  & 50.45  & & 46.70  \\
    \bottomrule
}

\newcommand{\TableAblationQualityInput}{
    \toprule
    
     & & & \multicolumn{2}{c}{R1} & & mAP \\
    \cline{4-5} \cline{7-7}    
    \vspace{-0.38cm}$\mathbf{C}_p$  & \vspace{-0.37cm}$\mathbf{C}_s$ & \vspace{-0.38cm}$\mathbf{C}_e$ & @0.5 & @0.7 & & Avg. \\
    \midrule
        \cmark &  &  & 62.97  & 49.03  & & 46.26  \\
         & \cmark &  & 64.39  & 50.71  & & 46.66  \\
         &  & \cmark & 64.58  & 49.23  & & 46.71  \\
        \rowcolor{defaultcolor}
        \cmark & \cmark & \cmark & 65.10  & 51.61  & & 47.61 &  \\
    \bottomrule
}

\newcommand{\TableAblationAttn}{
    \toprule
    
    Anchor & Boundary & \multicolumn{2}{c}{R1} & & mAP \\
    \cline{3-4} \cline{6-6}    
    query  & query & @0.5 & @0.7 & & Avg. & & \vspace{-0.38cm}FLOPs & & \vspace{-0.38cm}Params\\
    \midrule
        \multicolumn{2}{c}{Global (shared)} & 62.26  & 49.03  & & 44.18 & & 0.71G & & 8.2M \\
        Global & Global & 63.61  & 49.81  & & 44.54 & & 0.72G & & 11.5M \\
        \rowcolor{defaultcolor}
        Global & Focused & 63.74  & 49.68  & & 46.21  & & 0.65G &  & 9.5M & \\
        Focused & Focused & 62.48  & 49.16  & & 45.87  & & 0.62G & & 7.6M \\
    \bottomrule
}

\begin{table*}[t]
    \caption{Ablative experiments on QVHighlights. Default settings are marked \colorbox{defaultcolor}{gray}.
    }
    \centering
    \subfloat[
    Number of predictions
    \label{tab:num_queries}
    ]{
    \centering
    \begin{minipage}[!b]{0.28\linewidth}{
    \begin{center}
    \tablestyle{1pt}{1.1}
    \scriptsize
    {
    \begin{tabularx}{0.95\linewidth}{p{0.7cm}<{\centering}p{0.7cm}<{\centering}p{0.7cm}<{\centering}p{0.2mm}<{\centering}p{0.7cm}<{\centering}p{0.1mm}}
        \TableAblationNumQueries
    \end{tabularx}
    }
    \end{center}
    }
    \end{minipage}
    }
    \subfloat[
    \scriptsize
    Choice of memory features
    \label{tab:boundary_features}
    ]{
    \begin{minipage}[!b]{0.34\linewidth}{
    \begin{center}
    \tablestyle{1pt}{1.1}
    \scriptsize
    {
    \begin{tabularx}{0.95\linewidth}{p{1.4cm}<{\centering}p{0.7cm}<{\centering}p{0.7cm}<{\centering}p{0.2mm}<{\centering}p{0.7cm}<{\centering}p{0.1mm}}
        \TableAblationBoundaryFeat
    \end{tabularx}
    }
    \end{center}
    }
    \end{minipage}
    }
    \subfloat[
    \scriptsize
    Number of boundary points
    \label{tab:num_boundary_points}
    ]{
    \begin{minipage}[!b]{0.33\linewidth}{
    \begin{center}
    \tablestyle{1pt}{1.1}
    \scriptsize
    {
    \begin{tabularx}{0.95\linewidth}{p{1.3cm}<{\centering}p{0.7cm}<{\centering}p{0.7cm}<{\centering}p{0.2mm}<{\centering}p{0.7cm}<{\centering}p{0.1mm}}
        \TableAblationNumBoundaryPoint
    \end{tabularx}
    }
    \end{center}
    }
    \end{minipage}
    }\\
    \vspace{0.5em}
       \subfloat[
    \scriptsize
    Query choices for quality prediction
    \label{tab:quality_input}
    ]{
    \centering
    \begin{minipage}[!b]{0.39\linewidth}{
    \begin{center}
    \tablestyle{1pt}{1.1}
    \scriptsize
    {
    \begin{tabularx}{0.96\linewidth}{p{0.6cm}<{\centering}p{0.6cm}<{\centering}p{0.6cm}<{\centering}p{0.7cm}<{\centering}p{0.7cm}<{\centering}p{0.2mm}<{\centering}p{0.7cm}<{\centering}p{0.1mm}}
        \TableAblationQualityInput
    \end{tabularx}
    }
    \end{center}
    }
    \end{minipage}
    }
    \subfloat[
    \scriptsize
    Combinations of attention layers for updating pathways
    \label{tab:attn_combination}
    ]{
    \centering
    \begin{minipage}[b]{0.59\linewidth}{
    \begin{center}
    \tablestyle{1pt}{1.1}
    \scriptsize
    {
    \begin{tabularx}{0.98\linewidth}{p{1.25cm}<{\centering}p{1.25cm}<{\centering}p{0.7cm}<{\centering}p{0.7cm}<{\centering}p{0.2mm}<{\centering}p{0.7cm}<{\centering}p{0.2mm}<{\centering}p{0.83cm}<{\centering}p{0.2mm}<{\centering}p{0.8cm}<{\centering}p{0.1mm}}
        \TableAblationAttn
    \end{tabularx}
    }
    \end{center}
    }
    \end{minipage}
    }
    \\
    \vspace{1.5em}
    
    \label{tab:ablations}
    \vspace{-1.0em}
\end{table*}
\begin{figure*}[t]
    \centering
    \begin{minipage}[b]{0.3\linewidth} 
        \centering
        \includegraphics[width=\linewidth]{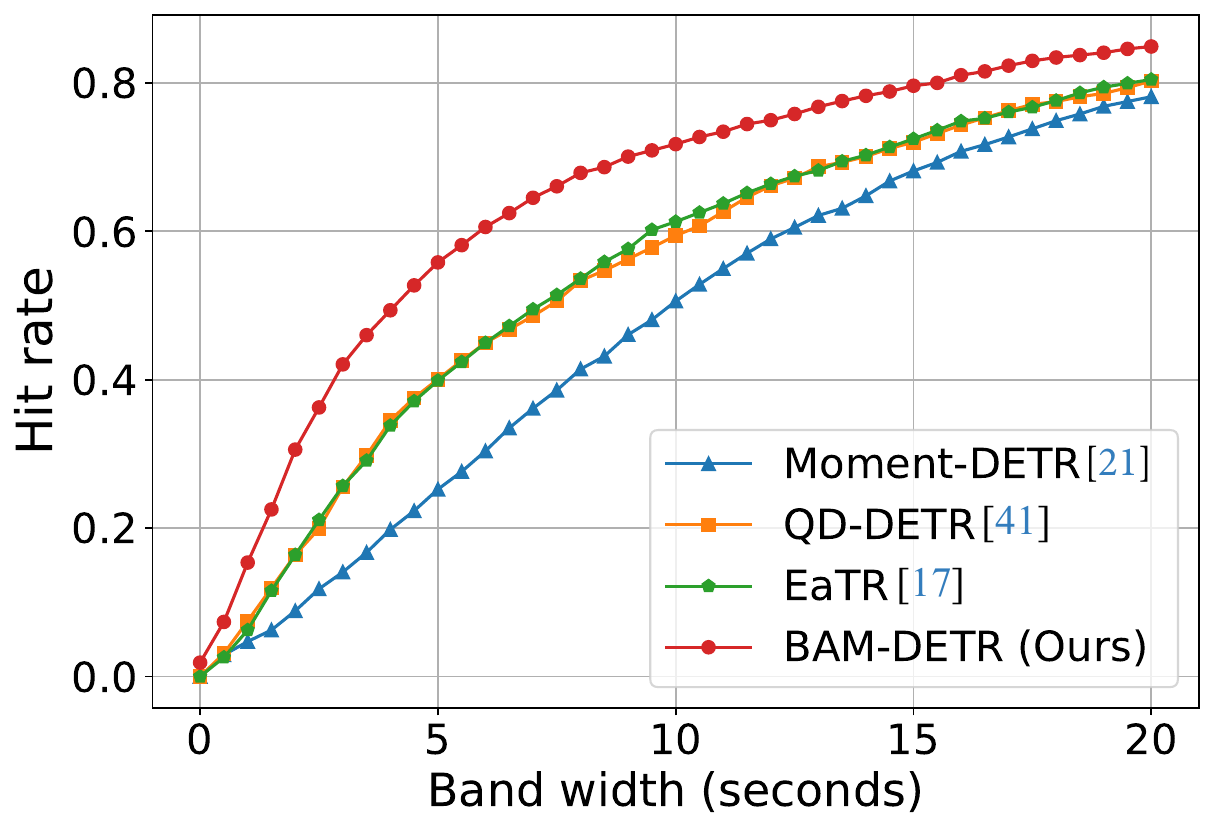}
        \subcaption{{\scriptsize Boundary hit rate}}
        \label{fig:boundary_hit_rate}
    \end{minipage}
    \hfill 
    \begin{minipage}[b]{0.66\linewidth} 
        \centering
        \includegraphics[width=\linewidth]{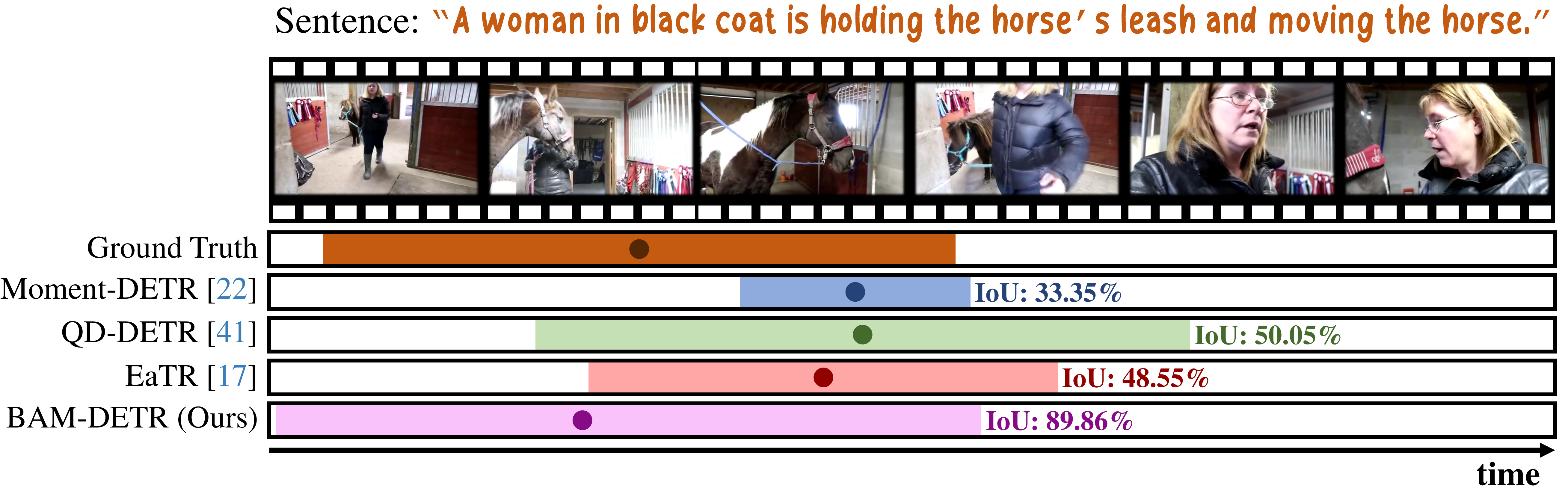}
        \subcaption{{\scriptsize Visualization results}}
        \label{fig:qualitative_results}
    \end{minipage}
    \caption{Analytical experiments on QVHighlights.}
    \label{fig:combined}
\end{figure*}

\subsection{Analysis}
We conduct analytical experiments on the QVHighlights validation split.

\noindent\textbf{Effect of each component.}~
We analyze the effect of each component in \Cref{tab:component_ablation}.
The direct adoption of our moment modeling solely improves the performance, particularly at strict thresholds, showcasing its advantage in boundary alignment.
Employing the dual-pathway decoder leads to further gains, which suggests the essential role of separate pathways.
Lastly, the quality-based scoring considerably elevates the scores, especially in terms of mAPs.

\noindent\textbf{Number of predictions.}
We experiment with varying numbers of predictions in \Cref{tab:num_queries}, where the model achieves robust results when $M$ is sufficiently large.
By default, we set $M$ to 10 for a fair comparison with the previous works~\cite{lei2021qvhighlights,moon2023qd-detr}.

\noindent\textbf{Locality-enhanced features.}
We analyze the effect of the choice of input features for boundary-focused attention in \Cref{tab:boundary_features}.
The results indicate that boundary-sensitive features are slightly more helpful than plain ones in precise localization.
In addition, the merged features achieve the best performance.

\noindent\textbf{Number of sampled boundary points.}~
We analyze the effect of the number of sampled boundary points $K$ in \Cref{tab:num_boundary_points}.
For comparisons, we report the case where the fixed boundary features are sampled without using offsets (1$^{\text{st}}$ row).
As shown in the table, the dynamic selection of neighborhoods rather than fixed boundaries is important for accurate moment localization.
The performance improves when sampling multiple points, while it saturates at $K=3$.

\noindent\textbf{Query choices for quality prediction.}~
Our model leverages both anchor and boundary queries for quality prediction.
We investigate the effect of query choices in \Cref{tab:quality_input}.
As a result, utilizing all the queries shows better localization performance than using one of them.
This might be trivial as different queries contain complementary information for quality prediction.

\noindent\textbf{Attention layers.}~
\Cref{tab:attn_combination} compares different combinations of attention layers for updating pathways in terms of performances and costs.
For an apple-to-apple comparison, we leverage the plain memory features for both attention layers.
First of all, employing separate global cross-attention for different queries leads to a huge parameter increase yet limited score gains.
Replacing the global one for boundary queries with our focused attention layer substantially reduces the cost, while achieving the best performance.
Meanwhile, the use of focused attention for both types of queries leads to inferior results.

\noindent\textbf{Boundary alignment.}~Standard IoU-based metrics are indirect measures for boundary alignment.
To precisely diagnose the ability, inspired by Deeplab~\cite{chen2017deeplab}, we compute the boundary hit rate of predictions with varying band widths.
We expand ground-truth boundary points with a band width to form starting/ending zones and regard a prediction as correct if both of its boundaries fall within the corresponding zones.
More details can be found in Appendix.
To disentangle the effect of ranking, we mark a video as correct if at least one prediction is correct and measure the video-level hit rate.
As shown in \cref{fig:boundary_hit_rate}, our model greatly outperforms the recent competitors.
The sharp increase in low band widths clearly validates the superiority of our model in boundary alignment.

\noindent\textbf{Qualitative results.}~As shown in \cref{fig:qualitative_results}, previous models fail to localize accurate moments with misleading center predictions.
Especially, QD-DETR~\cite{moon2023qd-detr} accurately predicts the moment length but suffers from the misaligned center, leading to the limited IoU.
In contrast, thanks to the novel moment modeling, our model predicts well-aligned boundaries without relying on center prediction.

\section{Conclusion}
\label{sec:conclusion}

In this paper, we identified the center misalignment issue of existing query-based models for sentence grounding.
To address it, we presented boundary-oriented moment modeling where boundaries are directly predicted without relying on centers.
Based on the modeling, we designed the boundary-aligned moment detection transformer characterized by dual-pathway decoding.
Further, we proposed localization quality-based scoring of predictions.
The efficacy of the proposed methods is validated by thorough examinations.
We hope this work sheds light on the issue of center-based moment modeling in detection transformers.

\section*{Acknowledgements}
This project was supported by the National Research Foundation of Korea grant funded by the Korea government (MSIT) (No. 2022R1A2B5B02001467; RS-2024-00346364).


\section{Formulation of Saliency Losses}
\label{sec:formulation_saliency}

As mentioned in the main paper, we adopt saliency losses for effective multimodal alignment in the encoder as in the common practice~\cite{lei2021qvhighlights,moon2023qd-detr}.
In specific, our total saliency-based loss is composed of three losses, \ie, $\mathcal{L}_{\text{sal}} = \mathcal{L}_{\text{margin}} + \mathcal{L}_{\text{cont}} + \mathcal{L}_{\text{neg}}$.
The margin-based loss $\mathcal{L}_{\text{margin}}$, defined in \cref{eq:loss_margin} of the main paper, aims to encourage the model to produce higher saliency scores for the clips relevant to the given sentence compared to less related clips.
Meanwhile, the rank-aware contrastive loss $\mathcal{L}_{\text{cont}}$ is utilized to preserve the ground-truth clip ranking in predicted saliency scores.
To be concrete, we first define the positive and negative sets based on an arbitrary reference score $r$, \ie, clips whose saliency score labels are higher than $r$ belongs to the positive set $\mathcal{B}_{r}^{+}$, and the remaining clips constitute the negative set $\mathcal{B}_{r}^{-}$.
The rank-aware contrastive loss is then formulated using a set of reference scores $\mathcal{R}$ as follows.
\begin{equation}
    \mathcal{L}_{\text{cont}} = - \sum_{\forall r \in \mathcal{R}} \log \frac{\sum_{\forall \hat{v} \in \mathcal{B}_{r}^{+}} \text{exp}(S(\hat{v}) / \tau)}{\sum_{\forall \hat{v} \in (\mathcal{B}_{r}^{+} \cup \mathcal{B}_{r}^{-})} \text{exp}(S(\hat{v}) / \tau)},
\end{equation}
where $S(\cdot)$ is a learnable saliency score predictor and $\tau$ is a temperature (set to 0.5).
We define $\mathcal{R}$ to be the set of saliency score labels of positive clips within ground-truth moments.

The negative relation loss is based on the assumption that all video clips should exhibit low saliency scores when paired with unmatched (negative) sentences.
Formally, the loss can be defined as follows.
\begin{equation}
    \mathcal{L}_{\text{neg}} = - \sum_{\forall \hat{v}^{\text{neg}} \in \hat{\mathcal{V}}^{\text{neg}}}  \log(1 - S(\hat{v}^{\text{neg}})),
\end{equation}
where $\hat{\mathcal{V}}^{\text{neg}}$ denotes the memory features obtained by processing the video with a negative sentence through the encoder.
In our implementation, a negative sentence is sampled from a different video-sentence pair in the mini-batch.

\section{Comparison with Deformable DETR}
\label{sec:comparison_deformable}

The proposed boundary-focused attention layer incorporates deformable attention, which is first proposed in Deformable DETR~\cite{zhu2021deformable-detr}.
It was originally designed for computationally efficient global attention with multi-scale features in object detection.
In contrast, we employ deformable attention for local aggregation of neighbor features, aiding precise boundary prediction in temporal sentence grounding.
To elucidate the discrepancy in their roles, we conduct a comparative experiment.
For this study, we implement a 1D variant of single-scale Deformable DETR, tailored for temporal sentence grounding.
We apply the proposed quality-based scoring to this model for a fair comparison.
Its key differences with our BAM-DETR lie in the moment formulation (center-based \textit{vs.} boundary-oriented) and the design of decoding layers (single-pathway \textit{vs.} dual-pathway).

\begin{table}[t]
\centering
\caption{Results on the QVHighlights validation split.}
\vspace{-2mm}
\scriptsize
\setlength{\tabcolsep}{0pt}
\begin{tabularx}{0.71\linewidth}{@{\hspace{2mm}}p{3.3cm}p{1cm}<{\centering}p{1cm}<{\centering}p{1.0mm}<{\centering}p{1cm}<{\centering}p{1cm}<{\centering}p{1cm}<{\centering}p{1.0mm}}
    \toprule    
     & \multicolumn{2}{c}{R1} & & \multicolumn{3}{c}{mAP} \\
    \cmidrule{2-3} \cmidrule{5-7}    
    \vspace{-0.45cm}Method    &@0.5 &@0.7 & &@0.5 &@0.75 &Avg. \\
    \midrule
    Deformable DETR & 60.52  & 49.35  & & 62.53 & 46.41 & 44.73  \\
    BAM-DETR (Ours) & 65.10  & 51.61  & & 65.41 & 48.56 & 47.61  \\
    \bottomrule
\end{tabularx}
\label{tab:deform_comparison}
\end{table}
\begin{figure}[t]
    \centering
    \includegraphics[width=0.52\linewidth]{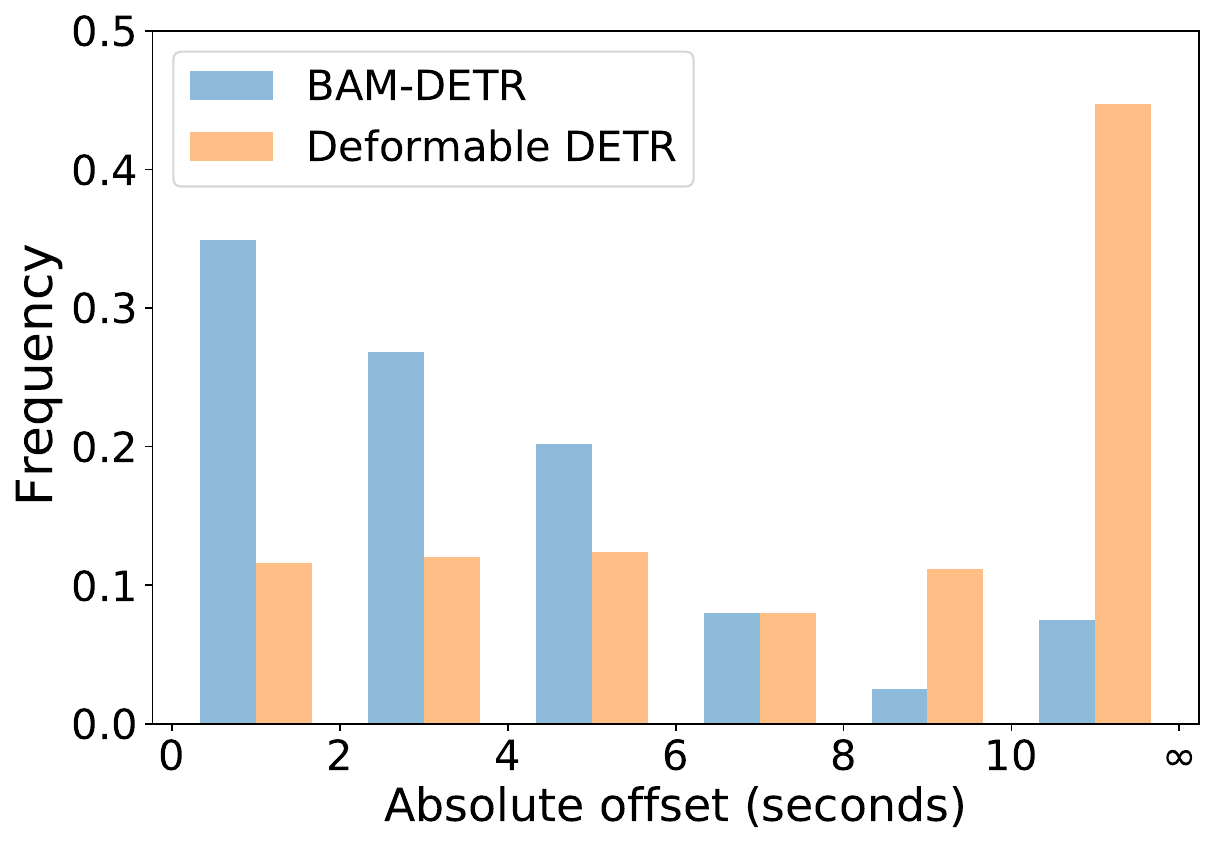}
    \vspace{-2mm}
    \caption{Offset histogram on the QVHighlights validation split.}
    \label{fig:offset_histogram}
\end{figure}

To analyze the behavior of deformable attention, we look into the absolute values of predicted offsets, \ie, how distant features are referenced during the attention process.
These offsets can indicate whether the attention is responsible for global or local interaction.
\cref{fig:offset_histogram} presents a visual comparison between the normalized histograms of predicted offsets from two comparative methods.
In our BAM-DETR, the deformable attention primarily concentrates on the neighbor features near the boundaries, \eg, over 80\% of offsets are shorter than 5 seconds.
Conversely, in the case of Deformable DETR, the deformable attention strives to aggregate global information, \eg, about 45\% of offsets are longer than 10 seconds.
These results clearly confirm the different roles of deformable attention in the two models.
In addition, we compare the grounding performance in \Cref{tab:deform_comparison}, where our BAM-DETR substantially outperforms Deformable-DETR.
This underscores the importance of our boundary-oriented moment modeling as well as the design of dual-pathway decoding layers.

\section{Details of Boundary Alignment Evaluation}
\label{sec:boundary_alignment_detail}
We provide more details regarding the experimental setup of boundary alignment evaluation performed in \cref{fig:boundary_hit_rate} of the main paper.
Inspired by the trimap evaluation of DeepLab~\cite{chen2017deeplab}, we propose a novel metric of boundary hit rate under varying band widths to evaluate the degree of boundary alignment.
In detail, we expand boundary points of the $n$-th ground truth $\{t_{s_n}, t_{e_n}\}$ with a given \textit{band width} of $l_w$ to form boundary zones.
We can denote the starting and ending zones by $Z_{s_n} = [t_{s_n} - 0.5l_w, t_{s_n} + 0.5l_w]$ and $Z_{e_n} = [t_{e_n} - 0.5l_w, t_{e_n} + 0.5l_w]$, respectively.
Then, for the $m$-th proposal $\{\hat{t}_{s_m}, \hat{t}_{e_m}\}$, we check whether both of its boundaries fall in the corresponding zones.
We iterate this process for all combinations of ground truths and predictions, and mark a video as correct if any pair is positive.
Formally, the binary variable of $h$ of a video is defined as:
\begin{equation*}
\begin{aligned}
    h =& \max_{\forall n, m} \Big[ \text{Hit}^s(n, m) \cdot \text{Hit}^e(n, m) \Big], \\
    \text{where}~~\text{Hit}^z(n, m) &= \mathbbm{1}\big[ |\hat{t}_{z_m} - t_{z_n}| \leq 0.5l_w \big],~~z \in \{s, e\}. \\
\end{aligned}
\end{equation*}
Note that we measure the hit rate over the whole validation set.

\section{More Analyses}
\label{sec:analyses}

\begin{table}[t]
\centering
\caption{Ablation study on the loss functions on QVHighlights.}
\vspace{-2mm}
\scriptsize
\setlength{\tabcolsep}{0pt}
\begin{tabularx}{0.8\linewidth}{p{0.9cm}<{\centering}p{0.9cm}<{\centering}p{0.9cm}<{\centering}p{0.9cm}<{\centering}p{0.9cm}<{\centering}p{1.0mm}<{\centering}p{1.0cm}<{\centering}p{1.0cm}<{\centering}p{1.0mm}<{\centering}p{1.0cm}<{\centering}p{1.0cm}<{\centering}p{1.0cm}<{\centering}p{1.0mm}}
\toprule
 & & & & & & \multicolumn{2}{c}{R1} & & & \multicolumn{3}{c}{mAP} \\
\cmidrule{7-8} \cmidrule{10-12}    
\vspace{-0.45cm}$\mathcal{L}_{\text{loc}}$  & \vspace{-0.45cm}$\mathcal{L}_{\text{cls}}$ & \vspace{-0.45cm}$\mathcal{L}_{\text{qual}}$ & \vspace{-0.45cm}$\mathcal{L}_{\text{sal}}$ & \vspace{-0.45cm}$\mathcal{L}_{\text{regul}}$ & & @0.5 & @0.7 & & @0.5 & @0.75 & Avg. \\
\midrule
    \cmark & \cmark &  &  & & & 56.77  & 41.03  & & 58.63 & 39.25 & 39.12  \\
    \cmark & \cmark &  & \cmark &  & & 60.58  & 46.65  & & 62.09 & 43.83 & 42.94  \\
    \cmark & \cmark &  & \cmark & \cmark  & & 63.61  & 50.26  & & 63.01 & 44.98 & 44.16 \\
\midrule
    \cmark &  & \cmark &  &  & & 59.23  & 46.13  & & 60.52 & 44.80 & 43.48  \\
    \cmark &  & \cmark & \cmark &  & & 63.23  & 50.00  & & 64.03 & 47.42 & 46.64  \\
    \rowcolor{defaultcolor}
    \cmark &  & \cmark & \cmark & \cmark  & & 65.10 & 51.61 & & 65.41 & 48.56 & 47.61  \\
\bottomrule
\end{tabularx}
\label{tab:loss_ablation}
\end{table}

\noindent\textbf{Ablation study on loss functions.}~Our model employs several loss functions for training.
We conduct an ablative experiment to diagnose their effects.
\Cref{tab:loss_ablation} summarizes the results, where the upper part adopts the typical classification-based scoring whereas the lower one leverages our proposed quality-based scoring.
We first examine the benefit of saliency losses.
Consistent with the recent findings~\cite{lei2021qvhighlights}, we observe that the saliency losses effectively guide the cross-modal alignment in the encoder, leading to notable performance improvements.
Then we investigate the importance of our regularization loss designed for boundary-sensitive feature construction  (\cf, \cref{eq:loss_regul} of the main paper).
It can be observed that regardless of the choice of scoring methods, the boundary regularization leads to significant performance boosts.
Putting together the results in \Cref{tab:boundary_features} of the main paper, it becomes clear that boundary-sensitive features are essential for precise boundary updating.
Lastly, the comparison between the two separate parts validates the efficacy of our quality-based scoring, especially in terms of mAPs.

\begin{figure}[t]
    \centering
    \begin{minipage}[b]{0.35\linewidth} 
        \centering
        \includegraphics[width=\linewidth]{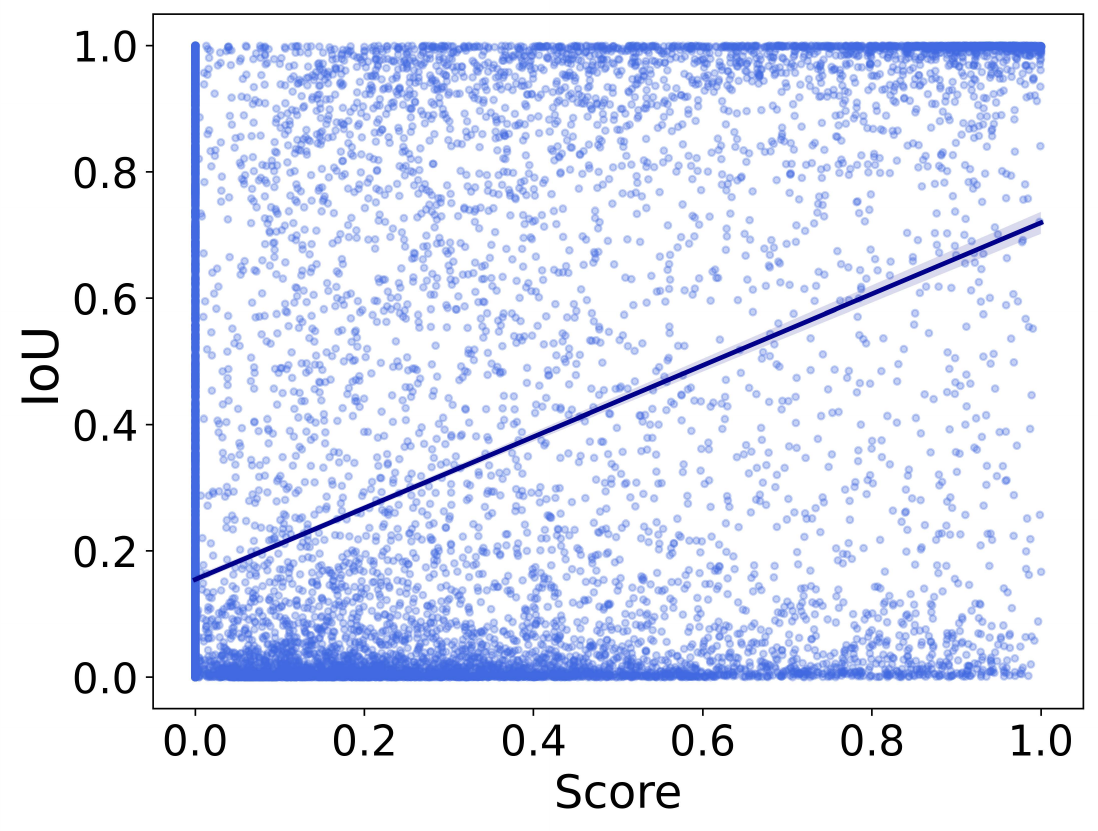}
        \subcaption{Classification-based scoring}
        \label{fig:cor_class}
    \end{minipage}
    \hspace{1.5em}
    \begin{minipage}[b]{0.35\linewidth} 
        \centering
        \includegraphics[width=\linewidth]{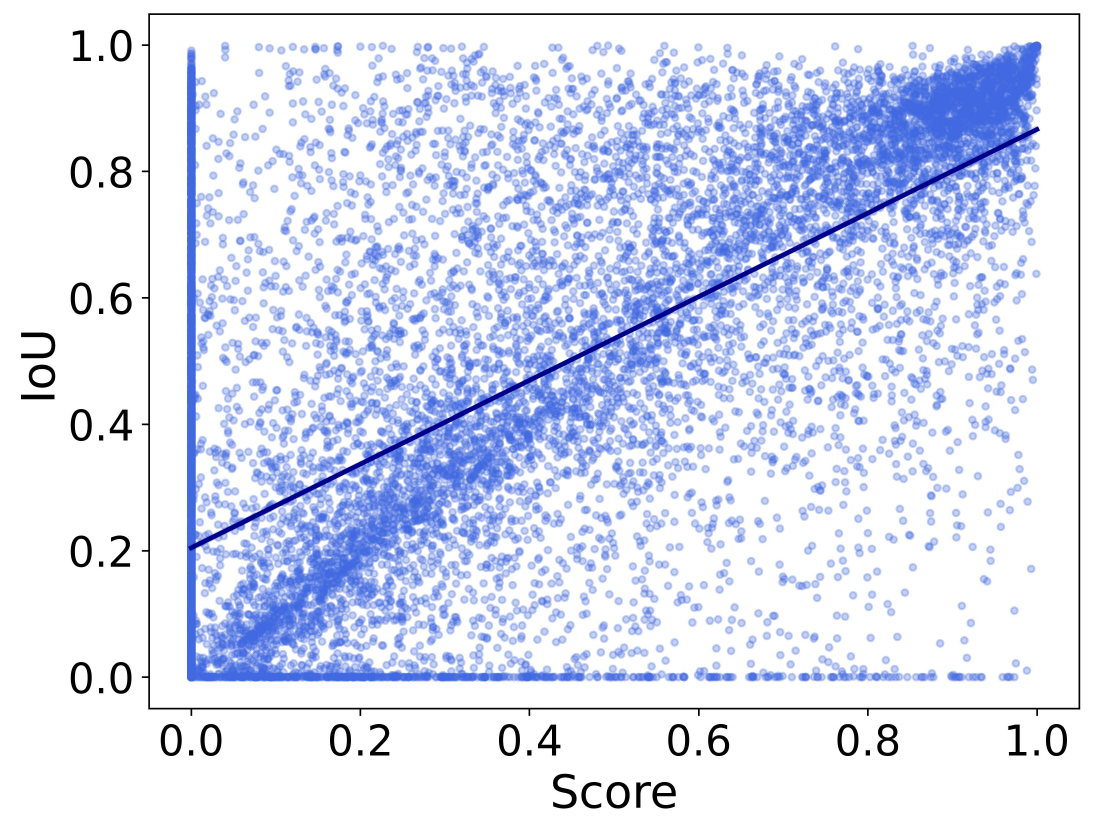}
        \subcaption{Quality-based scoring}
        \label{fig:cor_quality}
    \end{minipage}
    \vspace{-2mm}
    \caption{Correlation between scores and IoUs with ground truths: (a) the classification scores show a moderate correlation (Pearson's r of 0.44); (b) the quality scores exhibit a stronger correlation (Pearson's r of 0.67).}
    \label{fig:correlation_figure}
\end{figure}

\noindent\textbf{Comparison between scoring methods.}~We present the quality-based scoring method to replace the conventional classification-based one.
To compare two scoring methods, we draw scatter plots of scores \textit{vs.} IoUs with ground truths using all predictions on the QVHighlights validation set.
\cref{fig:cor_class} shows that classification scores correlate with IoUs to an extent.
On the other hand, we observe in \cref{fig:cor_quality} that our quality-based scoring shows a much stronger correlation with IoUs.
These results validate its efficacy in estimating the localization qualities of proposals, indicating that it is more appropriate for proposal ranking.

\noindent\textbf{Generalizablity of the quality-based scoring.}~By design, our quality-based scoring method is generalizable to any query-based approach.
To investigate this property, we conduct experiments by adopting the quality-based scoring on top of three representative models: Moment-DETR~\cite{lei2021qvhighlights}, QD-DETR~\cite{moon2023qd-detr}, and EaTR~\cite{jang2023eatr}.
The results are shown in \Cref{tab:qualtiy_generalization}, where the proposed scoring method brings consistent improvements over different baselines.
Noticeably, we can observe the pronounced gains at high IoU thresholds, which indicates better alignment of proposals with the ground truths.
This corroborates our claim that moment proposals ought to be ranked based on their localization qualities rather than the degree of matching.

\begin{table}[t]
\centering
\caption{Generalizability evaluation of quality-based scoring on the QVHighlights validation split.}
\vspace{-2mm}
\scriptsize
\setlength{\tabcolsep}{0pt}
\begin{threeparttable}
\begin{tabularx}{0.735\linewidth}{@{\hspace{2mm}}p{3.6cm}p{1.0cm}<{\centering}p{1.0cm}<{\centering}p{1.0mm}<{\centering}p{1.0cm}<{\centering}p{1.0cm}<{\centering}p{1.0cm}<{\centering}p{1.0mm}}
    \toprule    
     & \multicolumn{2}{c}{R1} & & \multicolumn{3}{c}{mAP} \\
    \cmidrule{2-3} \cmidrule{5-7}    
    \vspace{-0.45cm}Method    &@0.5 &@0.7 & &@0.5 &@0.75 &Avg. \\
    \midrule
    Moment-DETR\tnote{\ddag}~\cite{lei2021qvhighlights} & 53.23  & 34.00  & & 54.80 & 29.02 & 30.58  \\
    \hspace{1mm}$+$ quality-based scoring & 56.77  & 38.65  & & 55.09 & 35.30 & 34.98  \\
    \midrule
    QD-DETR\tnote{\ddag}~\cite{moon2023qd-detr} & 62.90  & 46.77  & & 62.66 & 41.51 & 41.24  \\
    \hspace{1mm}$+$ quality-based scoring & 64.26  & 50.32  & & 63.79 & 46.03 & 44.50  \\
    \midrule
    EaTR\tnote{\ddag}~\cite{jang2023eatr} & 57.74  & 42.71  & & 59.40 & 39.34 & 39.06  \\
    \hspace{1mm}$+$ quality-based scoring & 59.42  & 45.61  & & 60.24 & 42.29 & 41.61  \\
    \bottomrule
\end{tabularx}
\vspace{0.5mm}
\hspace{0.3mm}
\tnote{\ddag}All models are reproduced by official codebase
\end{threeparttable}
\label{tab:qualtiy_generalization}
\end{table}

\noindent\textbf{Efficiency comparison.}~We perform an efficiency comparison with previous state-of-the-art methods in terms of computational costs (\# of FLOPs) and memory (\# of Parameters).
The comparison results on the QVHighlights validation set are shown in \Cref{tab:efficiency_comparison}.
We can observe that our BAM-DETR has a comparable model size with EaTR~\cite{jang2023eatr}.
In terms of localization performance, it outperforms all the existing approaches by large margins, especially under strict evaluation metrics, which is consistent with the test split results (\cf, \Cref{tab:qvhighlights} of the main paper).
To make a fairer comparison, we also implement a small variant of our model equipped with slimmer encoding layers, namely BAM-DETR$^{slim}$.
In detail, we halve the hidden dimension of the encoder and reduce the number of fully-connected layers within each attention block.
As a result, BAM-DETR$^{slim}$ can achieve better efficiency with a cost of slightly sacrificing localization performance.
Nevertheless, it is shown that BAM-DETR$^{slim}$ suffices to largely surpass the existing approaches even with fewer parameters and FLOPs.
These results confirm the effectiveness of the proposed method.

\begin{table}[t]
\centering
\caption{Efficiency comparison results on the QVHighlights validation split.}
\vspace{-2mm}
\scriptsize
\setlength{\tabcolsep}{0pt}
\begin{threeparttable}
\begin{tabularx}{0.85\linewidth}{@{\hspace{2mm}}p{2.5cm}p{1.0cm}<{\centering}p{1.0cm}<{\centering}p{1.0mm}<{\centering}p{1.0cm}<{\centering}p{1.0cm}<{\centering}p{1.0cm}<{\centering}p{2.0mm}<{\centering}p{1.0cm}<{\centering}p{2.0mm}<{\centering}p{1.0cm}<{\centering}p{1.0mm}}
    \toprule    
     & \multicolumn{2}{c}{R1} & & \multicolumn{3}{c}{mAP} \\
    \cmidrule{2-3} \cmidrule{5-7}    
\vspace{-0.4cm}
    Method & @0.5 & @0.7 & & @0.5 & @0.75 & Avg. & & \vspace{-0.45cm}FLOPs & & \vspace{-0.45cm}Params\\
    \midrule
        QD-DETR$^\ddagger$~\cite{moon2023qd-detr} & 62.90  & 46.77  & & 62.66 & 41.51 & 41.24 & & 0.59G & & 7.7M \\
        UniVTG$^{\dagger\ddagger}$~\cite{lin2023univtg} & 59.74  & 40.90  & & 58.61 & 36.76 & 36.13  & & 0.98G &  & 43.4M \\
        EaTR$^\ddagger$~\cite{jang2023eatr} & 60.90  & 46.13  & & 62.01 & 42.17 & 41.43 & & 0.47G & & 9.1M \\
        \midrule
        BAM-DETR$^{slim}$ & 63.94  & 50.19  & & 64.51 & 48.51 & 47.03  & & 0.43G &  & 7.2M \\
        BAM-DETR & 65.10  & 51.61  & & 65.41 & 48.56 & 47.61  & & 0.65G &  & 9.5M \\
    \bottomrule
\end{tabularx}
\vspace{0.5mm}
\tnote{~~\dag}The hidden dimension is four times larger than that of competitors \\
\tnote{~~\ddag}All models are reproduced by official checkpoints
\end{threeparttable}
\label{tab:efficiency_comparison}
\end{table}

\section{Further Qualitative Results}
\label{sec:qualitative_results}

\begin{figure*}[t]
    \centering
    \includegraphics[width=\linewidth]{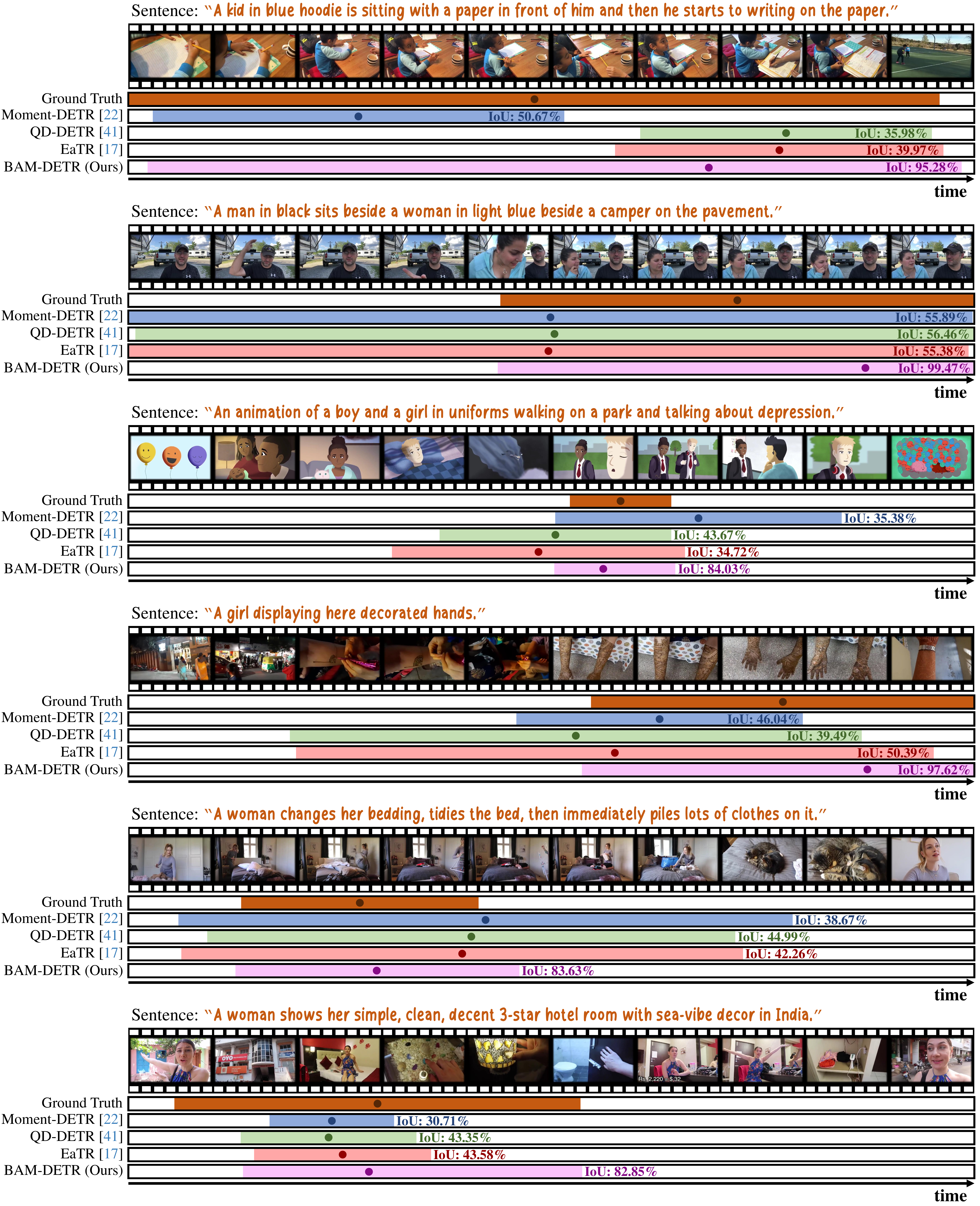}
    \vspace{-6mm}
    \caption{Qualitative comparison on the QVHighlights validation split.}
    \label{fig:qualitative_suppl}
\end{figure*}

\begin{figure*}[t]
    \centering
    \includegraphics[width=\linewidth]{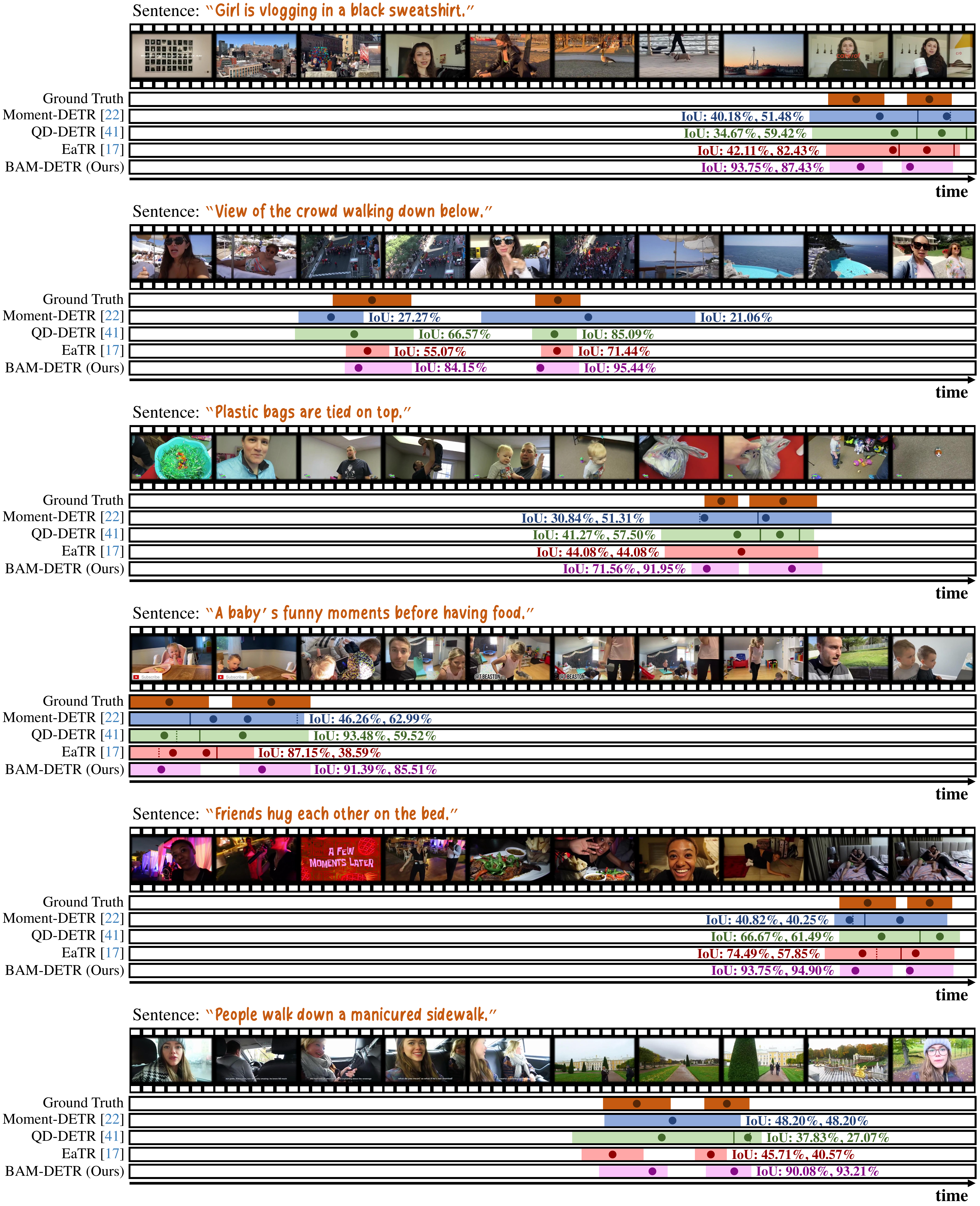}
    \vspace{-6mm}
    \caption{Qualitative comparison on the QVHighlights validation split.}
    \label{fig:qualitative_suppl_2}
\end{figure*}

We perform further qualitative comparisons with previous query-based methods in \cref{fig:qualitative_suppl} and \cref{fig:qualitative_suppl_2}.
The comparison results across various scenarios demonstrate the superiority of our BAM-DETR over the strong competitors.

\clearpage

%
%
\bibliographystyle{splncs04}
\bibliography{main}
\end{document}